\documentclass[journal]{IEEEtran}
\usepackage{lineno,hyperref}
\hypersetup{hidelinks,
	colorlinks=true,
	allcolors=black,
	pdfstartview=Fit,
	breaklinks=true}
\usepackage{amsthm}
\usepackage{bbm}
\usepackage{bm}
\usepackage{amsfonts}
\usepackage{amsmath}
\usepackage{mathrsfs}
\usepackage{amssymb}
\usepackage{enumerate}
\usepackage{graphicx}
\usepackage{subfigure}
\usepackage{booktabs}
\usepackage{graphicx}
\usepackage{setspace}
\usepackage{algorithm}
\usepackage{algorithmic}
\usepackage{setspace}
\usepackage{multirow}
\usepackage{color}
\usepackage{cite}
\usepackage{array}
\usepackage{times}
\usepackage{mathptmx}
\usepackage{dsfont}
\usepackage{xcolor}
\usepackage{url}
\ifCLASSINFOpdf
\else
\fi
\begin{document}
\title{Single-Image HDR Reconstruction Assisted Ghost Suppression and Detail Preservation Network for Multi-Exposure HDR Imaging}
\author{Huafeng~Li, Zhenmei~Yang, Yafei~Zhang, Dapeng~Tao, Zhengtao~Yu % <-this % stops a space
\thanks{This work was supported in part by the National Natural Science Foundation of China under Grant 62161015, Grant 62276120, the Yunnan Fundamental Research Projects (202301AV070004). (\textit{Huafeng Li and Zhenmei Yang contributed equally to this work.}) (\textit{Corresponding author: Yafei Zhang.})} 
\thanks{H. Li, Z. Yang, Y. Zhang and Z. Yu are with the Faculty of Information Engineering and Automation, Kunming University of Science and Technology, Kunming 650500, China.(E-mail:lhfchina99@kust.edu.cn (H. Li); yangTongxue123\_km@163.com(Z. Yang); zyfeimail@163.com(Y. Zhang); ztyu@hotmail.com(Z. Yu))}
\thanks{D. Tao is with FIST LAB, School of Information Science and Engineering, Yunnan University, Kunming 650091, China.(E-mail:dapeng.tao@gmail.com)}% <-this % stops a space
%\thanks{G. Qi is affiliated with Computer Information Systems Department, State University of New York at Buffalo State, Buffalo, NY 14222, USA. (E-mail:qig@buffalostate.edu)}
\thanks{}%J. Luo is with the Department of Computer Science, University of Rochester, Rochester, NY 14623 (J. Luo, e-mail:jiebo.luo@gmail.com).}% <-this % stops a space
%\thanks{ ${\ast}$ indicates contributed to this work equally.}%co-first %authors.}
\thanks{Manuscript received xxxx;}}
\markboth{Journal of \LaTeX\ Class Files}%
{Shell \MakeLowercase{\textit{et al.}}}
\maketitle
\begin{abstract}
The reconstruction of high dynamic range (HDR) images from multi-exposure low dynamic range (LDR) images in dynamic scenes presents significant challenges, especially in preserving and restoring information in oversaturated regions and avoiding ghosting artifacts. While current methods often struggle to address these challenges, our work aims to bridge this gap by developing a multi-exposure HDR image reconstruction network for dynamic scenes, complemented by single-frame HDR image reconstruction. This network, comprising single-frame HDR reconstruction with enhanced stop image (SHDR-ESI) and SHDR-ESI-assisted multi-exposure HDR reconstruction (SHDR-A-MHDR), effectively leverages the ghost-free characteristic of single-frame HDR reconstruction and the detail-enhancing capability of ESI in oversaturated areas. Specifically, SHDR-ESI innovatively integrates single-frame HDR reconstruction with the utilization of ESI. This integration not only optimizes the single image HDR reconstruction process but also effectively guides the synthesis of multi-exposure HDR images in SHDR-A-MHDR. In this method, the single-frame HDR reconstruction is specifically applied to reduce potential ghosting effects in multi-exposure HDR synthesis, while the use of ESI images assists in enhancing the detail information in the HDR synthesis process.  Technically, SHDR-ESI incorporates a detail enhancement mechanism, which includes a self-representation module and a mutual-representation module, designed to aggregate crucial information from both reference image and ESI. To fully leverage the complementary information from non-reference images, a feature interaction fusion module is integrated within SHDR-A-MHDR. Additionally, a ghost suppression module, guided by the ghost-free results of SHDR-ESI, is employed to suppress the ghosting artifacts. Experimental results on four public datasets demonstrate the efficacy and superiority of the proposed method. The code is available at \url{https://github.com/lhf12278/SAMHDR}.

\end{abstract}
\begin{IEEEkeywords}
Single-image HDR reconstruction, dynamic scene HDR imaging, enhancement stop image, self-representation, mutual-representation.
\end{IEEEkeywords}
\IEEEpeerreviewmaketitle
\section{Introduction}
High dynamic range (HDR) imaging technology excels in the portrayal of real-world images, offering a wider range of brightness. This characteristic endows it widely applicable in virtual reality, traffic surveillance, digital television, medical imaging, and other fields. However, ordinary digital cameras are constrained by their limited dynamic imaging range. The captured images not only fall far short of the visible brightness range to the human eye but also suffer from the loss of detail information in overexposed regions. In the early stages of HDR imaging, the acquisition of HDR images relied on high-end cameras and dedicated sensors\cite{Guicquero2016AnAlgorithm, Gnanasambandam2020HDR, Miandji2021Compressive}. However, the prohibitive cost of this approach posed a substantial barrier to the widespread utilization of HDR imagery. If it were feasible to synthesize or reconstruct HDR images from low dynamic range (LDR) images captured by ordinary cameras, the reliance on high-end cameras can be effectively eliminated. Therefore, investigating the methodology for reconstructing HDR images from LDR counterparts is the key to promoting the development of HDR imaging.

\begin{figure}[t!]
\centering
\includegraphics[width=2.7in,height=1.4in]{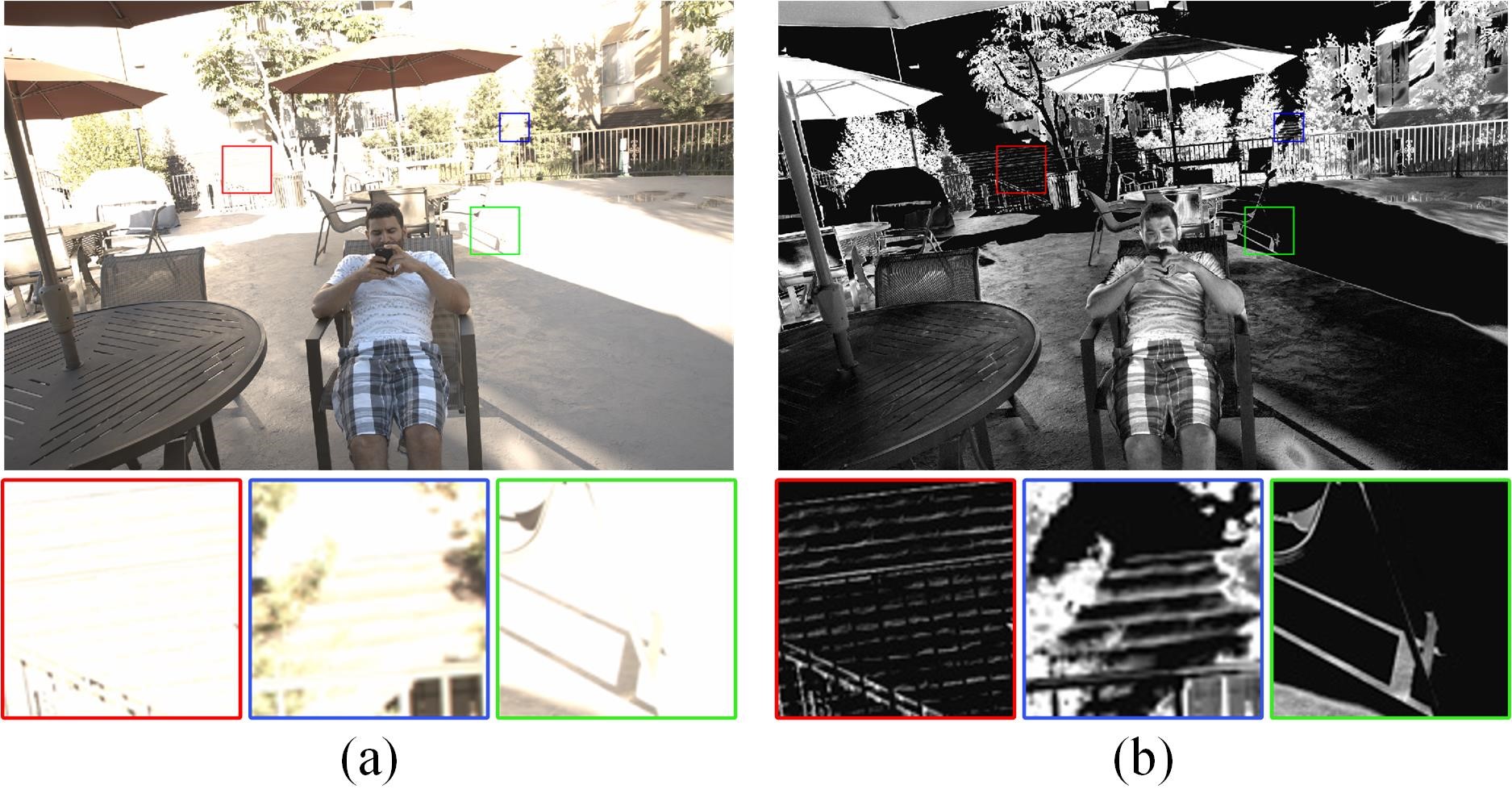}
\caption{(a) Reference image. (b) Enhancement stop image (ESI). ESI has the capability to highlight subtle information in oversaturated regions of the reference image. }
\label{label1}
\end{figure}

Existing HDR imaging techniques can classified into two categories: single-frame HDR image reconstruction (SHDR)  \cite{khan2019fhdr, akyuz2020deep, chen2021single, dinh2023end} and multi-exposure HDR synthesis \cite{Prabhakar2021Self-Gated,catley2022flexhdr, yan2022high, tang2023structure}. SHDR typically reconstructs the HDR image from a given single LDR image. Most of traditional SHDR methods \cite{akyuz2007hdr,banterle2007framework,huo2014physiological,kovaleski2014high} only focus on expanding the dynamic range of LDR image, resulting in poor quality of the reconstructed HDR image. Currently, deep learning based SHDR techniques \cite{ning2018learning,chen2021hdrunet,she2023single,Eilertsen-2017-HDR-CNNS} have garnered significant attention in the research community. Compared to traditional methods, deep learning based approaches have improved the visual quality of reconstruction results. However, owing to the non-uniform exposure inherent in LDR images, there exist regions with missing information, posing a challenge for SHDR methods to achieve visually satisfactory results. This challenge is particularly pronounced in overexposed areas with severe information loss. In contrast to SHDR, multi-exposure HDR synthesis methods can achieve superior reconstruction results due to their ability to integrate complementary information derived from different LDR images.

However, in dynamic scenes, multi-exposure HDR synthesis often faces the risk of introducing ghosting artifacts, which affects the visual quality of the reconstruction results. Around ghost suppression, a series of effective methods have been proposed. Those methods primarily eliminate ghosting induced by target movement through either explicit alignment \cite{Bogoni2000extending,hu2013hdr,kang2003high,zimmer2011freehand,prabhakar2019fast,Kalantari-2017-Deep-scenes} or implicit alignment\cite{Yan-2019-AHDRNet,Deng-2021-Selective,Chen-2022-APFusion,Yan-2020-Non-local,Pu-2020-robust-parallax,Liu-2021-ADNet,Tan-2022-high-saturation} of input LDR images. In explicit alignment methods, the common alignment strategy is image alignment based on optical flow \cite{baker2011database}. However, the performance of such methods heavily relies on the accuracy of optical flow estimation. When input images contain occluded or overexposed regions, the performance of optical flow estimation algorithms is adversely affected. This heightens the risk of introducing ghosting artifacts into HDR images. Compared to explicit alignment, implicit alignment methods can avoid the challenges posed by optical flow estimation. Typically, implicit alignment is carried out at the level of image features. Common implicit alignment methods mainly include attention-based methods\cite{Yan-2019-AHDRNet,Deng-2021-Selective,Chen-2022-APFusion,Yan-2020-Non-local} and deformable convolution-based methods\cite{Pu-2020-robust-parallax,Liu-2021-ADNet,Tan-2022-high-saturation}. Attention-based methods mitigate ghosting by highlighting crucial information and reducing the impact of ghosting-related data in HDR reconstruction. In contrast, deformable convolution-based methods suppress ghosting by aligning input image features. However, these methods exhibit limited performance in preserving valuable information and recovering lost data in oversaturated regions, thereby limiting the quality improvement of reconstructed HDR images. Although literature \cite{Niu-2021-HDR-GAN} attempts to recover lost information by generative adversarial network (GAN), the poor quality of generated content results in unsatisfactory reconstruction results. 

To address the issues of multi-exposure HDR synthesis, this paper proposes an end-to-end dual-branch directional promotion network. This network consists of a single-frame HDR image reconstruction with an embedded enhancement stop image (SHDR-ESI) and an SHDR-ESI-assisted multi-exposure HDR image reconstruction (SHDR-A-MHDR). SHDR-ESI primarily serves to aid SHDR-A-MHDR in suppressing ghosting artifacts within the fused features.
Owing to the precise spatial alignment between the reference image and ESI fed into the SHDR-ESI branch, the HDR reconstruction results for the single-frame HDR image are free from ghosting artifacts. However, as SHDR-A-MHDR takes LDR images with different exposures as inputs, its reconstruction results may contain ghosting artifacts.
Based on this fact, it is proposed to use the information difference between SHDR-ESI and SHDR-A-MHDR results to suppress non-shared information (ghosting artifacts) and preserve shared information. Specifically, the single-frame HDR reconstruction is applied to reduce potential ghosting effects in multi-exposure HDR synthesis, while the use of ESI assists in enhancing the detail information in the HDR synthesis process. As shown in Fig. \ref{label1}, ESI plays a pivotal role in highlighting subtle information in oversaturated regions. Therefore, the difficulty of information recovery in these regions will be effectively reduced with the assistance of ESI. 
To effectively utilize the information contained in ESI and the existing information in the reference image, a detail enhancement mechanism (DEM), consisting of a self-representation module (SRM) and a mutual-representation module (MRM), is designed in SHDR-ESI. SRM emphasizes shared information in the reference image and ESI that is beneficial for reconstruction quality. Moreover, the MRM is employed to transfer detailed information from ESI to the reference image, facilitating the aggregation of crucial information in both ESI and the reference image. To fully exploit complementary information in non-reference images, a feature interaction fusion module (FIFM) is designed in SHDR-A-MHDR. This module utilizes weights generated through the interaction of features to selectively combine information from input images. Simultaneously, it enhances the role of valuable information in HDR image reconstruction. Additionally, to further eliminate ghosting in the fused features, a ghost suppression module (GSM) guided by intermediate features from SHDR-ESI is introduced.
The GSM not only serves to suppress ghosting artifacts but also embeds the information from ESI into the reconstructed features. This achieves comprehensive preservation and restoration of features in oversaturated regions of reference images. In summary, the main contributions of this paper are as follows:
\begin{itemize}
\item An end-to-end dual-branch directional promotion network is proposed to achieve multi-exposure HDR image reconstruction and ghosting suppression. This network consists of two core components: SHDR-ESI and SHDR-A-MHDR. These components work in a unidirectional promotion manner to effectively suppress ghosting artifacts.

\item In SHDR-ESI, SRM and MRM are designed to effectively highlight and aggregate important information from ESI and the reference image, thus preventing the loss of subtle information in oversaturated regions. Simultaneously, FIFM and GSM are constructed in SHDR-A-MHDR. These modules emphasize the role of important information in HDR image reconstruction and suppress the influence of ghosting artifacts on fusion results while integrating features from multi-exposure LDR images.

\item The proposed method is evaluated through experiments on four challenging datasets and compared with state-of-the-art methods. Extensive experimental results demonstrate its effectiveness and superiority. In addition, ablation experiments are conducted on the core components of the proposed method to demonstrate their effectiveness.
\end{itemize}

\section{Related Work}
\subsection{Single-frame HDR Reconstruction}
The HDR reconstruction result obtained from a single image remains unaffected by ghosting. However, the task is more challenging due to the absence of supplementary information from oversaturated or underexposed areas. For the HDR image reconstruction from a single-frame LDR image, existing methods primarily fall into two categories. The first category involves direct recovery of HDR image from single-frame LDR image. Conversely, the second category encompasses the mapping of a single-frame LDR image to multi-exposure LDR images, which is subsequently followed by HDR image reconstruction based on these multi-exposure images.

For the recovery of HDR image from a single LDR image, Eilertsen $et \ al$. \cite{Eilertsen-2017-HDR-CNNS} proposed using a convolutional neural network (CNN) to predict missing information in overexposed areas and subsequently reconstructing the HDR image based on this prediction. Marnerides $et \ al$. \cite{Marnerides-2018-Expandnet} introduced a network called ExpandNet. To improve the quality of HDR image, this network adopts a multi-scale structural framework to avoid the introduction of upsampling layers. Yang $et \ al$. \cite{Yang-2018-Image-transformation} introduced an end-to-end deep reciprocating HDR transformation. This transformation includes two CNN networks, one for HDR image detail reconstruction and another for LDR image detail correction. Liu $et \ al$. \cite{Liu-2020-Single-pipeline} embedded domain knowledge related to LDR images into HDR image reconstruction model. Furthermore, they modeled the process of HDR-to-LDR image formation and employed three dedicated CNNs to reverse this process, thereby achieving HDR image reconstruction. The aforementioned single-frame LDR-to-HDR image recovery approaches have indeed proven to be effective. However, it still faces significant challenges in recovering missing information and preserving details in oversaturated and undersaturated regions, which limits the quality of the recovered images.

To mitigate the challenges faced in single-frame LDR-to-HDR image recovery, the approaches of mapping a single LDR image into multi-exposure LDR images and subsequently utilizing multi-exposure image fusion for HDR image reconstruction have been proposed\cite{Endo-2017-Deep-mapping,Lee-2018-Deep-networks,Kim-2021-End-to-end,Le-2023-Single-generation}. Specifically, Endo $et \ al$. \cite{Endo-2017-Deep-mapping} employed a supervised learning approach to transform a single-frame LDR image into multiple LDR images with different exposures. Then, the transformed results are fused to obtain an HDR image. Lee $et \ al$. \cite{Lee-2018-Deep-networks} utilized an adversarial generative network to transform an LDR image into multi-exposure stacks and consequently estimated the HDR image from them. To address the inversion artifacts in stack reconstruction-based methods, Kim $et \ al$. \cite{Kim-2021-End-to-end} proposed a fully differentiable HDR imaging technique. Le $et \ al$. \cite{Le-2023-Single-generation} introduced a weakly supervised learning approach that aims to reverse the physical formation process of HDR image by training a model to generate multiple exposure images from a single image. These SHDR methods not only reduce the steps of image feature alignment and tone mapping but also effectively prevent the introduction of ghosting artifacts. However, these methods still face significant challenges in recovering lost details in overexposed areas. In contrast, the HDR image reconstruction methods based on multi-exposure LDR images fusion can effectively address the issue of information loss caused by overexposure due to the inherent information complementarity among multiple images.

\begin{figure*}[t!]
\centering
\includegraphics[width=0.8\textwidth]{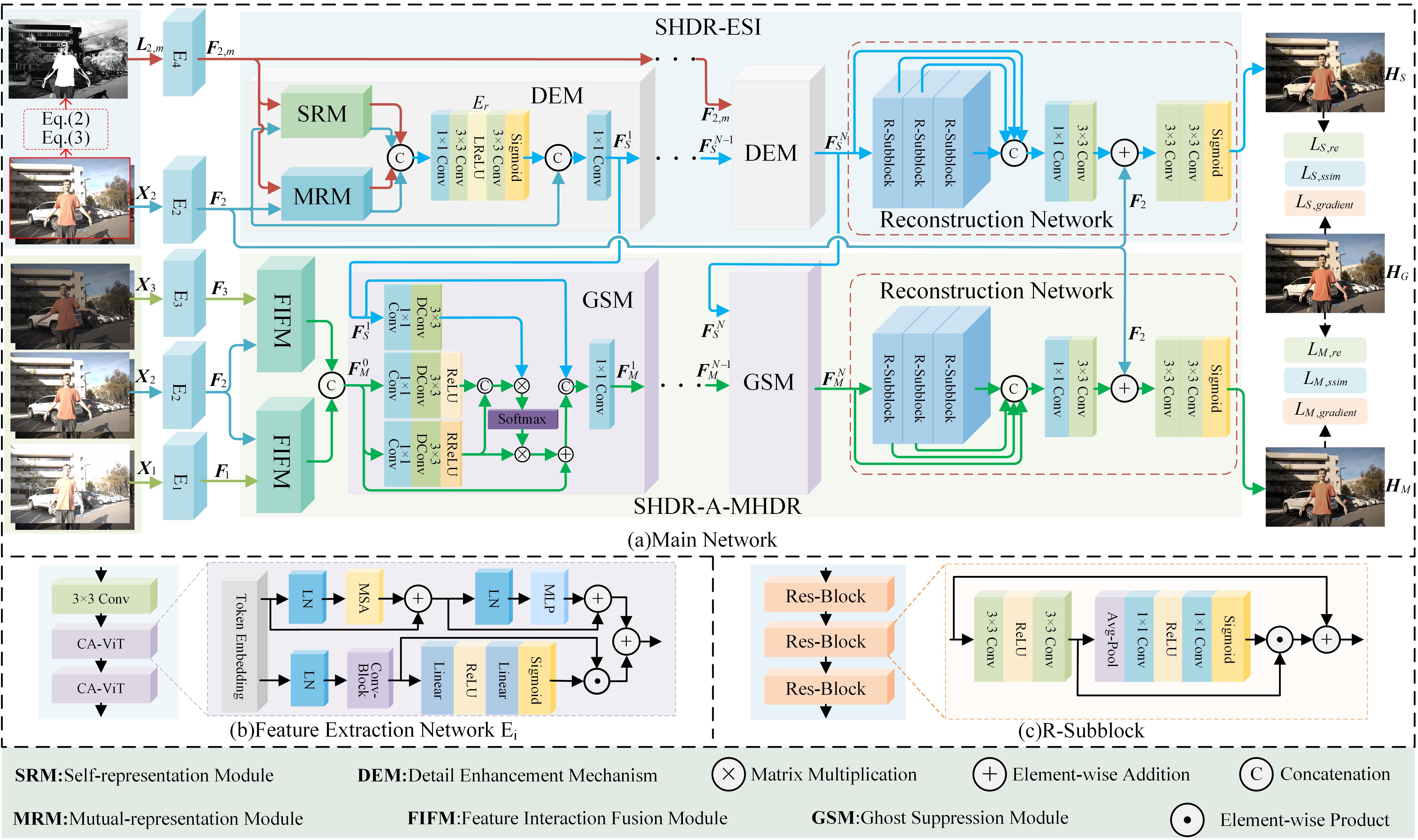}
\caption{Overall framework of the proposed method. (a) the main network consists of the SHDR-ESI branch and the SHDR-A-MHDR branch, where SHDR-ESI primarily assists SHDR-A-MHDR in suppressing ghosting artifacts within the fused features. SRM and MRM are employed to  highlight and aggregate valuable information from both the reference image and ESI, within the SHDR-ESI branch. In the SHDR-A-MHDR branch, FIFM is designed to merge the information from input images and enhance the role of valuable information in HDR image reconstruction. Subsequently, GSM is employed to mitigate potential ghosting artifacts within the fused features. (b) the feature extraction network $E_{i} (i=1,2,3,4)$ is composed of a $3\times 3$ convolution and two CA-ViT \cite{Liu-2022-Ghost-free}. (c) R-Subblock consists of three residual blocks (Res-blocks).}
\label{label2}
\end{figure*}

\subsection{Multi-Exposure HDR Image Reconstruction}
Multi-exposure LDR images encompass a wealth of complementary information, making them particularly valuable for addressing the challenges encountered in SHDR. This characteristic has attracted researchers' attention. However, reconstructing HDR images from multi-exposure LDR images with large-scale foreground object motion poses the risk of introducing ghosting artifacts. In the field of HDR reconstruction with ghosting artifact suppression, significant progress has been made in recent years. Currently, deep learning-based methods for multiple-exposure HDR reconstruction can be divided into CNN-based methods and hybrid CNN and Transformer methods.
\subsubsection{CNN-Based Methods}
Kalantari $et \ al$. \cite{Kalantari-2017-Deep-scenes} employed optical flow algorithms to align the input LDR images with a reference image. Subsequently, the aligned images are fed into a CNN for fusion to reconstruct the HDR image. Wu $et \ al$. \cite{Wu-2018-Deep-motions} introduced the first non-optical flow-based deep learning framework to address HDR imaging in dynamic scenes. This method transforms the HDR imaging problem into a non-optical flow conversion problem from LDR images to HDR images. The transformed and fused LDR images are reconstructed into HDR images using a CNN with an encoder-decoder architecture. Due to the ability of attention mechanisms to highlight useful information, Yan $et \ al$. \cite{Yan-2019-AHDRNet}, Deng $et \ al$. \cite{Deng-2021-Selective} and Chen $et \ al$. \cite{Chen-2022-APFusion} have proposed attention-based methods incorporated into the CNN framework to suppress ghost artifacts caused by large-scale foreground movement. Yan $et \ al$. \cite{Yan-2020-Non-local} proposed embedding a non-local operation into the intermediate layers of a U-net architecture to remove ghost artifacts introduced in HDR reconstruction by calculating non-local correlations within the input images. 
Given that deformable convolutions can dynamically adjust the pixels involved in the computation to achieve feature alignment, Pu $et \ al$. \cite{Pu-2020-robust-parallax} and Liu $et \ al$.\cite{Liu-2021-ADNet} have proposed HDR reconstruction methods based on deformable convolutions. Prabhakar $et \ al$. \cite{Prabhakar-2020-Towards} proposed a bilateral guided upsampler method for HDR reconstruction, which removes ghost in HDR images through motion estimation and motion compensation. Chung $et \ al$. \cite{Chung-2022-WACV} transformed the motion alignment problem into a simple brightness adjustment problem. They adjusted the brightness of reference image features using underexposed and overexposed image features to generate well-aligned multi-exposure features, which were then used to reconstruct HDR images.

The aforementioned methods primarily focus on suppressing ghost artifacts during the multi-exposure image fusion process, but tend to neglect the recovery of lost information in overexposed or underexposed regions. To address this issue, Niu $et \ al$. \cite{Niu-2021-HDR-GAN} employed a GAN to integrate features from multi-exposure images and recover information in overexposed regions. Although this method is somewhat effective, it may not focus extensively on ghost artifact suppression. Recently, Yan \textit{et al}.\cite{HDR-Diffusion} proposed an HDR reconstruction network based on conditional diffusion model \cite{DDPM}. They formulated the task of HDR deghosting as an image generation problem, leveraging LDR features as conditions for the diffusion model. This introduces a new perspective to the field of HDR image reconstruction. Nevertheless, a significant challenge remains in achieving complete suppression of ghosting within multi-exposure HDR synthesis. 

\subsubsection{Hybrid CNN and Transformer Methods}
Transformer architecture excels in capturing global image features more effectively. Consequently, HDR image reconstruction methods based on Transformer have been proposed. Song $et \ al$. \cite{Song-2022-TransHDR} proposed generating a mask for the ghost region from the input LDR images, dividing the given images into ghost and non-ghost regions, and adaptively selecting either CNN or Transformer for HDR reconstruction. Liu $et \ al$. \cite{Liu-2022-Ghost-free} utilized CNN to extract features from the input LDR images and attention mechanisms to suppress ghosting artifacts. Finally, a Swin-Transformer embedded with channel attention was used to reconstruct the HDR image. Yan $et \ al$. \cite{Yan-2023-unified} developed a network called HyHDRNet for reconstructing HDR images from multi-exposure LDR images. This network consists of a CNN-based content alignment sub-network and a Transformer-based fusion sub-network.  Subsequently, Yan $et \ al$. \cite{Yan-2023-SMAE} combined CNN and multi-scale Swin-Transformer residual blocks and introduced a two-stage training mode to achieve semi-supervised HDR imaging. Chen $et \ al$. \cite{Chen-2023-AAAI} employed deformable convolutions to align the LDR image features extracted by CNN and introduced Transformer in the fusion process to model long-range relationships between features. The methods mentioned above primarily aim to address the impact of ghosting on the HDR image. Certain approaches comprehensively address the issues of ghosting suppression and information loss in overexposed regions. However, these methods may not effectively address the preservation of useful information and the recovery of lost information in underexposed areas, simultaneously. This limitation hinders the potential for enhancing the visual quality of reconstruction results. In contrast to the aforementioned approaches, this paper introduces a novel paradigm by integrating ghost suppression, information preservation and recovery within a unified framework. It presents a distinctive method for multi-exposure HDR reconstruction in dynamic scenes guided by single-frame HDR reconstruction.

\section{The Proposed Method}
\subsection{Overview}
The overall structure of the proposed method is mainly composed of single-frame HDR reconstruction with ESI embedding (SHDR-ESI) and SHDR-ESI-assisted multi-exposure HDR reconstruction (SHDR-A-MHDR), as illustrated in Fig. \ref{label2}. SHDR-ESI is used to assist SHDR-A-MHDR in suppressing ghosting artifacts and  enhancing the information in overexposed regions. To ensure the high-quality HDR image produced by SHDR-ESI, we propose integrating ESI into SHDR process. Concurrently, a DEM, composed of an SRM and an MRM, is employed to highlight critical information and aggregate complementary feature effectively. Within the SHDR-A-MHDR framework, both reference and non-reference image features are fused through the FIFM. Furthermore, the intermediate features derived from SHDR-ESI are fed into the GSM to effectively suppress ghosting artifacts.

\subsection{Image Preprocessing}
Assuming there are three multi-exposure LDR source images, denoted as $\left\{ { \bm 
 L_{1},\bm L_{2},\bm L_{3}}\right\}$. A similar processing approach can be applied for additional source images. Typically, to achieve satisfactory reconstruction results, it is advisable to perform gamma correction on the source images, i.e., mapping the LDR images to HDR domain to obtain their corresponding HDR images $\left\{{\bm  H_{1},\bm H_{2},\bm H_{3}}\right\}$:
\begin{equation}
\begin{aligned}
    \bm H_{i}= \frac{\bm L_{i}^{\gamma}}{t_{i}},\left( i=1,2,3 \right)
\end{aligned}
\end{equation}
where $t_{i}$ denotes the exposure time of $\bm L_{i}$, and $\gamma$ is the gamma correction parameter, typically set to 2.2 based on empirical knowledge\cite{tang2023structure}. 

After the correction, $\bm L_{i}$ and $\bm H_{i}$ are concatenated on the channel dimension to obtain $\bm X_{i}=\left[\bm L_{i}, \bm H_{i} \right]$, which is then used as the input for the feature extraction network. In multi-exposure HDR reconstruction, a medium exposure image is usually selected as the reference image. The complementary information from other exposure images is integrated into the reference image for HDR reconstruction. In the SHDR-ESI, the reference image is also used as the base image for HDR reconstruction.

Owing to non-uniform exposure levels, it's worth noting that the reference image also contains oversaturated areas. In these regions, certain details may not be distinctly displayed, consequently impacting SHDR process. To address this issue, this paper introduces an image edge-detail correction method called enhancement stop. This process is described in Equations $\bm \left( 2 \right)$ and $\bm \left( 3 \right)$ as follows:
\begin{equation}
\begin{aligned}
    \bm L_{2,PT}\left( x,y \right)= \sqrt{\bm L_{2,P}\left( x,y \right)^2 + \bm L_{2,T}\left( x,y \right)^2}
\end{aligned}
\end{equation}
\begin{equation}
   \bm L_{2,m}\left( x,y \right)=
   \begin{cases}
   1, &~ \bm L_{2,PT}\left( x,y \right)\geq c\\
   \bm L_{2,PT}\left( x,y \right), &~ otherwise
\end{cases}
\end{equation}
where $\bm L_{2,P}\left( x,y \right)$ and $\bm L_{2,T}\left( x,y \right)$ respectively represent the values of channel P and channel T at position $\left( x,y \right)$ after $\bm L_{2}$ is transformed into the IPT color space. $\bm L_{2,m}$ is the enhancement stop image. $ c $ is a constant, commonly set to 1. For further details about the enhancement stop image, please refer to \cite{kwon2023multi}. 

\begin{figure}[t!]
\centering
\includegraphics[width=2.6in,height=1.8in]{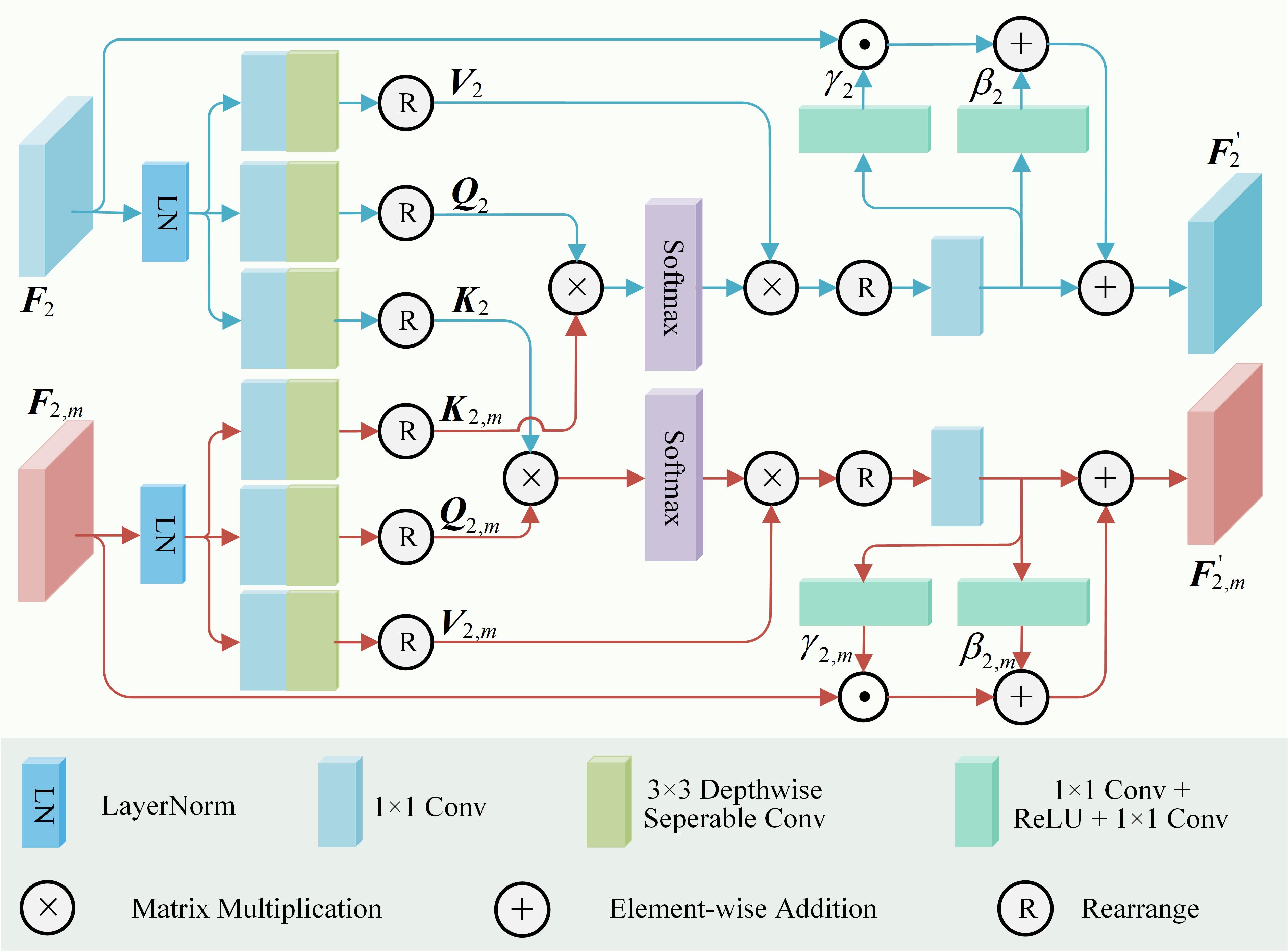}
\caption{Illustration of the self-representation module (SRM).}
\label{label3}
\end{figure}
\subsection{Single-Frame HDR Reconstruction with ESI Embedding}
The single-frame HDR image reconstruction branch in this study is designed to assist the multi-exposure HDR reconstruction branch in producing higher-quality HDR images. Typically, the reference image serves as the input for single-frame HDR image reconstruction. To improve the detail in over-saturated areas of the reference image, this paper proposes a single-frame HDR image reconstruction method embedded with ESI (SHDR-ESI). Furthermore, the DEM, comprising an SRM and an MRM, is utilized to recover and highlight the missing details and subtle information in overexposed areas, utilizing the detailed information provided by ESI.

\subsubsection{Self-representation Module}
The SRM primarily serves to emphasize the role of important information in HDR reconstruction. Its detailed structure is depicted in Fig. \ref{label3}. Let the reference image be $\bm X_{2}$ and the ESI be $\bm L_{2,m}$. Features $\bm F_{2} \in \mathbb R^{C\times H\times W}$ and $\bm F_{2,m} \in \mathbb R^{C\times H\times W}$ can be obtained after the images $\bm X_{2}$ and $\bm L_{2,m}$ pass through the feature extraction networks $E_{2}$ and $E_{4}$, respectively, which are designed based on CA-ViT\cite{Liu-2022-Ghost-free}. $\bm F_{2}$ and $\bm F_{2,m}$ are separately fed into the SRM. Following their input, they traverse through a normalization layer and subsequently undergo three parameter-unshared feature extraction blocks, each consisting of a $1\times 1$ convolutional layer and a $3\times 3$ depthwise separable convolutional layer. The outcomes are then rearranged to yield features $\bm Q_{2}$, $\bm K_{2}$, $\bm V_{2}$ and $\bm Q_{2,m}$, $\bm K_{2,m}$, $\bm V_{2,m}$. 

$\bm L_{2,m}$ carries a wealth of edge detail information from the source images, which corresponds to the salient detail information in $\bm X_{2}$. If $\bm L_{2,m}$ is employed to emphasize the detail information in $\bm X_{2}$, it will help to improve the quality of HDR reconstruction results. To this end, we propose a feature enhancement method with dynamic parameter modulation. As shown in Fig. \ref{label3}, this method initially utilizes $\bm F_{2}$ to find the corresponding features in $\bm F_{2,m}$ and subsequently enhances them as follows:
\begin{equation}
\begin{aligned}
    \bm F_{2}^{att}=Conv_{1\times 1}\bigg( Softmax\bigg( \frac{\bm Q_{2} (\bm K_{2,m})^{T}}{s_{2}} \bigg) \bm V_{2} \bigg)
\end{aligned}
\end{equation}
where $s_{2}$ is a learnable scale parameter, $Conv_{1\times 1}$ represents a $1\times 1$ convolutional layer. Similarly, we can use $\bm F_{2,m}$ to find the corresponding features in $\bm F_{2}$, and enhance the relevant features as follows:
\begin{equation}
\begin{aligned}
    \bm F_{2,m}^{att}=Conv_{1\times 1}\bigg( Softmax\bigg( \frac{\bm Q_{2,m} (\bm K_{2})^{T}}{s_{2,m}} \bigg) \bm V_{2,m} \bigg)
\end{aligned}
\end{equation}
where $s_{2,m}$ is a learnable scale parameter.

To prevent the information loss of input features, it is common practice to directly supplement the original features $\bm F_{2}$ and $\bm F_{2,m}$ into $\bm F_{2}^{att}$ and $\bm F_{2,m}^{att}$. Although this approach is effective, it falls short in effectively highlighting the role of key information in both $\bm F_{2}$ and $\bm F_{2,m}$ during HDR reconstruction. The feature enhancement method we propose, incorporating dynamic parameter modulation, offers a solution to this challenge. To obtain these dynamic modulation parameters $({{\bf{\gamma}}_2}, {{\bf{\beta }}_2})$ and $({{\bf{\gamma}}_{2,m}}, {{\bf{\beta }}_{2,m}})$, the paper suggests generating them based on features $\bm F_{2}^{att}$ and $\bm F_{2,m}^{att}$ by employing a combination of a $1\times 1$ convolutional layer, a ReLU activation function, and another $1\times 1$ convolutional layer.

The dynamic parameters $\left( \gamma_{2}, \beta_{2} \right)$ and $\left( \gamma_{2,m}, \beta_{2,m} \right)$ are used to modulate $\bm F_{2}$ and $\bm F_{2,m}$, respectively. The modulated features are individually added to $\bm F_{2}^{att}$ and $\bm F_{2,m}^{att}$ to obtain enhanced features $\bm F_{2}^{'}$ and $\bm F_{2,m}^{'}$:
\begin{equation}
\begin{aligned}
    \bm F_{2}^{'}=\bm F_{2}^{att} + \left( \bm F_{2} \odot \gamma_{2} + \beta_{2} \right)
\end{aligned}
\end{equation}
\begin{equation}
\begin{aligned}
    \bm F_{2,m}^{'}=\bm F_{2,m}^{att} + \left( \bm F_{2,m} \odot \gamma_{2,m} + \beta_{2,m} \right)
\end{aligned}
\end{equation}
where $\odot$ represents the Hadamard product. 

\begin{figure}[t!]
\centering
\includegraphics[width=2.6in,height=1.8in]{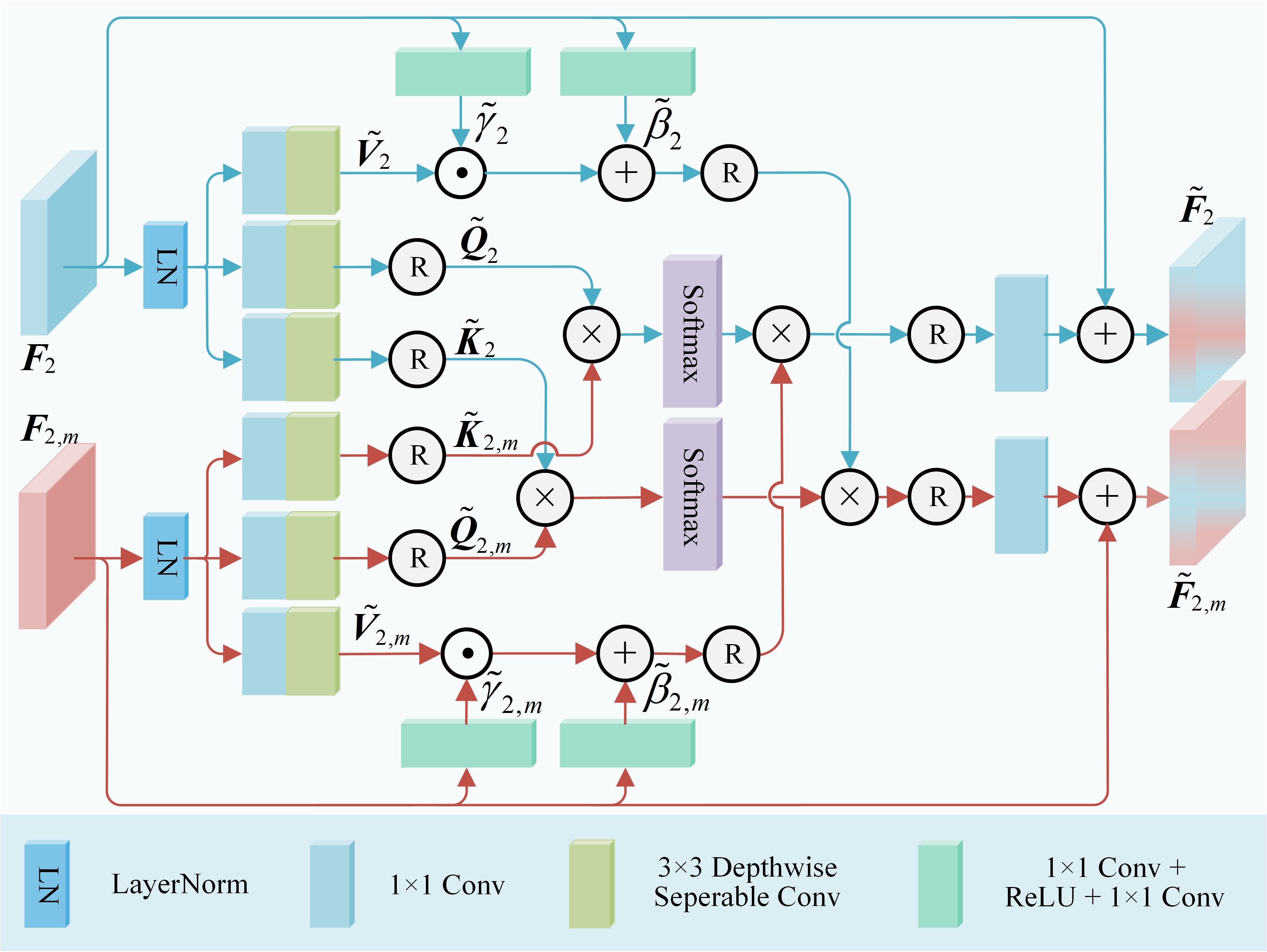}
\caption{Illustration of the mutual-representation module (MRM).}
\label{label4}
\end{figure}

\subsubsection{Mutual-representation Module}
The SRM is primarily designed to highlight the features relevant to the HDR reconstruction results. However, it does not possess the capability to transfer the information from ESI to the features of $\bm X_{2}$ to compensate for the loss of information in the overexposed areas of $\bm X_{2}$. To address this issue, this paper introduces an MRM, depicted in Fig. \ref{label4}, to facilitate the mutual transfer of information between $\bm X_{2}$ and $\bm L_{2,m}$. In this process, we utilize the features from $\bm X_{2}$ as one set of bases to represent $\bm L_{2,m}$, concurrently employing the features from $\bm L_{2,m}$ as another set of bases to represent $\bm X_{2}$, thereby achieving the mutual transfer of information.

The MRM, similar to the SRM, takes $\bm F_{2}$ and $\bm F_{2,m}$ as inputs. The feature extraction block of MRM is also composed of parallel layers, including LayerNorm, a $1\times 1$ convolutional layer, and a $3\times 3$ depth-wise separable convolutional layer. The results of $\bm F_{2}$ and $\bm F_{2,m}$ passing through three feature extraction blocks are denoted as $\tilde{\bm Q}_{2}$, $ \tilde{\bm K}_{2}$, $ \tilde{\bm V}_{2}$ and $ \tilde{\bm Q}_{2,m}$, $ \tilde{\bm K}_{2,m}$, $ \tilde{\bm V}_{2,m}$, respectively. 

To minimize information loss during the feature propagation process, this paper introduces a pair of modulation layers to modulate $\tilde{\bm V}_{2}$ and $\tilde{\bm V}_{2,m}$ with $\bm F_{2}$ and $\bm F_{2,m}$, respectively, before performing information exchange. This procedure aims to highlight the crucial information in $\tilde{\bm V}_{2}$ and $\tilde{\bm V}_{2,m}$. In order to achieve the interaction of information between $\bm F_{2}$ and $\bm F_{2,m}$, the mutual representation operation in this paper can be formulated as follows:
\begin{equation}
\begin{aligned}
    \tilde{\bm F}_{2}=Conv_{1\times 1}\bigg( Softmax\bigg( \frac{\tilde{\bm Q}_{2} (\tilde{\bm K}_{2,m})^{T}}{\tilde{s}_{2}} \bigg)\\
    \bigg(\tilde{\bm V}_{2,m}\odot \tilde{ \gamma}_{2} + \tilde{\beta}_{2} \bigg) \bigg) + \bm F_{2}
\end{aligned}
\end{equation}
\begin{equation}
\begin{aligned}
    \tilde{\bm F}_{2,m}=Conv_{1\times 1}\bigg( Softmax\bigg( \frac{\tilde{\bm Q}_{2,m} (\tilde{\bm K}_{2})^{T}}{\tilde{s}_{2,m}} \bigg)\\
    \bigg(\tilde{\bm V}_{2}\odot \tilde{ \gamma}_{2,m} + \tilde{\beta}_{2,m} \bigg) \bigg) + \bm F_{2,m}
\end{aligned}
\end{equation}
where modulation parameters $( \tilde{\gamma}_{2}, \tilde{\beta}_{2})$ and $( \tilde{\gamma}_{2,m}, \tilde{\beta}_{2,m})$ can be obtained from the following equations:
\begin{equation}
\begin{aligned}
    ( \tilde{\gamma}_{2}, \tilde{\beta}_{2})=Conv_{1\times 1}( ReLU( Conv_{1\times 1}( \bm F_{2} )))
\end{aligned}
\end{equation}
\begin{equation}
\begin{aligned}
    (\tilde{\gamma}_{2,m}, \tilde{\beta}_{2,m})=Conv_{1\times 1}( ReLU( Conv_{1\times 1}(\bm F_{2,m} )))
\end{aligned}
\end{equation}

\subsubsection{Reconstruction Network}
The features $\bm F_{2}^{'}$, $\bm F_{2,m}^{'}$, $\tilde{\bm F}_{2}$ and $\tilde{\bm F}_{2,m}$ are concatenated along the channel dimension and then processed through a network denoted as $E_{r}$. 
The outcome derived from this process is concatenated with $\bm F_{2}$ and then fed through a $1\times 1$ convolutional layer to obtain the aggregated features $\bm F_{s}^{1}$ of the first detail enhancement module (DEM). This process can be formulated as follows:
\begin{equation}
\begin{aligned}
    \bm F_{s}^{1}=Conv_{1\times 1}\big ( \big[E_{r} \big( \big[\bm F_{2}^{'},\bm F_{2,m}^{'},\tilde{\bm F}_{2},\tilde{\bm F}_{2,m} \big] \big),\bm F_{2} \big] \big)
\end{aligned}
\end{equation}
where $\left[\cdot,\cdot\right] $ denotes concatenation operation. $\bm F_{s}^{1}$ and $\bm F_{2,m}$ can be utilized as inputs for the second DEM to yield $\bm F_{s}^{2}$, and this process can be iteratively continued. Following $N$ DEM , the aggregate feature $\bm F_{s}^{N}$ is obtained.

Using $\bm F_{s}^{N}$ and $\bm F_{2}$ as the inputs of the reconstruction network, we obtain the final HDR image $\bm H_{s}$ of SHDR-ESI. As shown in Fig.\ref{label2}, the reconstruction network used in this paper is mainly composed of three R-Subblocks, each containing three residual blocks (Res-blocks).

\subsection{SHDR-ESI-assisted Multi-exposure HDR Reconstruction}
Because SHDR directly generates HDR images from the reference image, it can avoid the ghosting artifacts introduced by multi-exposure images fusion. In view of this characteristic of SHDR, this paper proposes SHDR-A-MHDR, which is intended to use the results of single-frame HDR reconstruction to suppress the ghosting artifacts in the results of multi-exposure HDR reconstruction. Specifically, FIFM is employed to merge the features from both the reference and non-reference images in a weighted manner. In addition, the GSM, which is guided by the ghost-free intermediate features derived from SHDR-ESI, is utilized to effectively suppress potential ghosting artifacts and enhance the valuable information within the fused features.

\begin{figure}[t!]
\centering
\includegraphics[width=2.8in,height=1.6in]{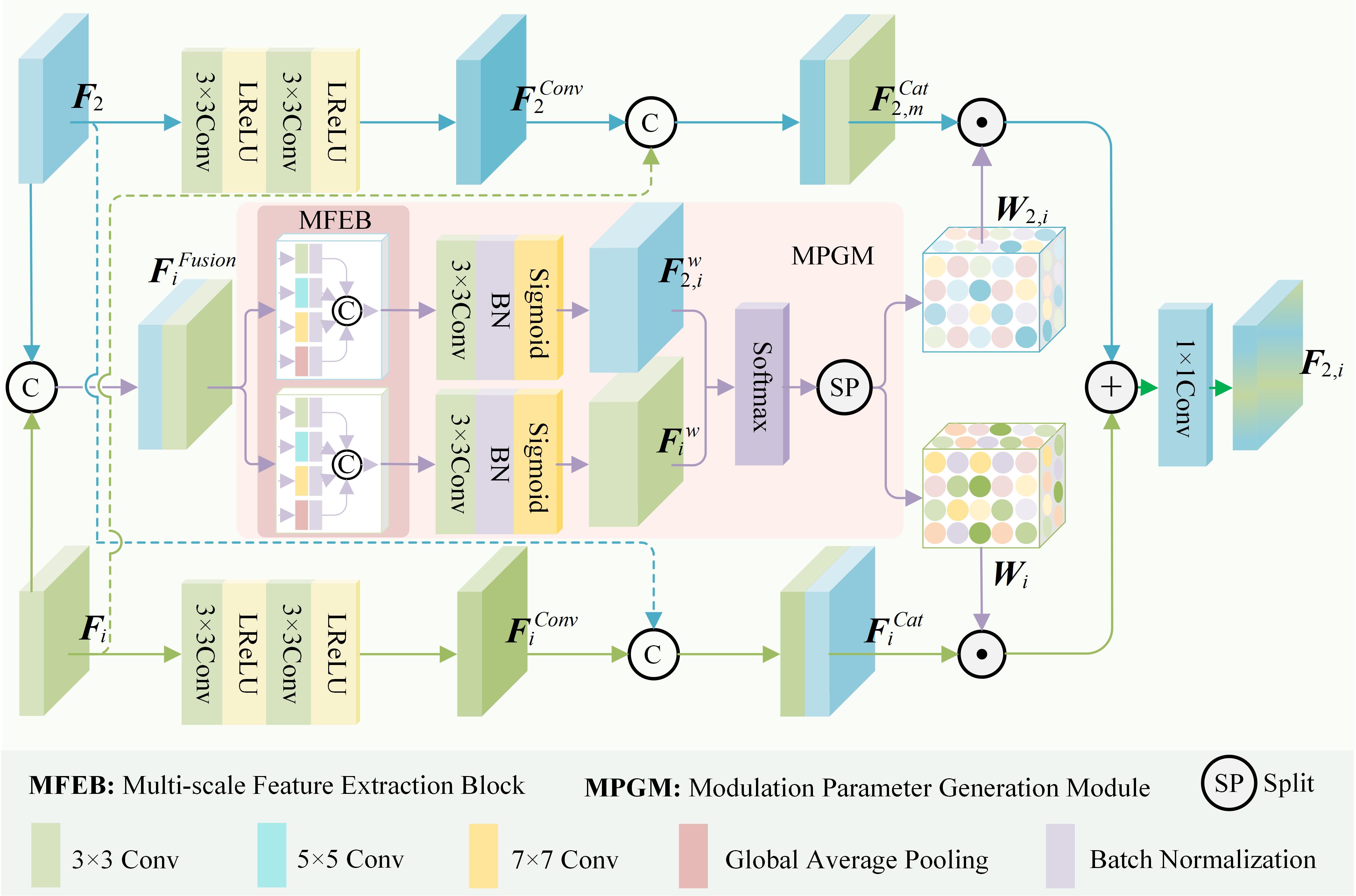}
\caption{Detailed structure of the feature interaction fusion module (FIFM).}
\label{label5}
\end{figure}

\subsubsection{Feature Interaction Fusion Module}
In HDR reconstruction, effectively highlighting the features that contribute to the improvement of reconstruction quality can enhance the quality of HDR image. To this end, this paper introduces an FIFM designed to facilitate information interaction between the reference image and other multi-exposure images. It enables the modulation of individual pixels in spatial positions based on the outcomes of this interaction. The specific structure of the FIFM is illustrated in Fig.\ref{label5}. To enhance the quality of HDR reconstruction image, this paper utilizes convolutional layers with a consistent structure to further extract features from both the reference feature $\bm F_{2}$ and the non-reference features $\bm F_{i}(i=1,3)$. These extracted features are represented as $\bm F_{2}^{Conv} \in \mathbb R^{C\times H\times W}$ and $\bm F_{i}^{Conv} \in \mathbb R^{C\times H\times W}(i=1,3)$:
\begin{equation}
\begin{aligned}
    \bm F_{2}^{Conv}=LReLU( Conv_{3\times 3}( LReLU( Conv_{3\times 3}\left( \bm F_{2} \right)) ) )
\end{aligned}
\end{equation}
\begin{equation}
\begin{aligned}
    \bm F_{i}^{Conv}=LReLU( Conv_{3\times 3}( LReLU( Conv_{3\times 3}\left( \bm F_{i} \right)) ) )
\end{aligned}
\end{equation}
where $LReLU$ and $Conv_{3\times 3}$ represent the LeakyReLU activation function and a $3\times 3$ convolution respectively. In order to more effectively utilize the complementary information from input features $\bm F_{2}$, $\bm F_{i}$ and features $\bm F_{2}^{Conv}$, $\bm F_{i}^{Conv}$, we cross-concatenate them to create $\bm F_{2,i}^{Cat}=\left[\bm F_{2} ,\bm F_{i}^{Conv} \right]$ and $\bm F_{i}^{Cat}=\left[\bm F_{i} ,\bm F_{2}^{Conv} \right]$ to promote information transfer between image features with different exposure. 

To independently modulate the feature at each spatial position within the interaction results $\bm F_{i}^{Cat}$ and $\bm F_{2,i}^{Cat}$, we have devised a modulation parameter generation module (MPGM). The input of the MPGM is represented as $\bm F_{i}^{Fusion}=\left[\bm F_{2} ,\bm F_{i} \right]$, which is obtained by concatenating the original features. In the MPGM, we design two sets of multi-scale feature extraction blocks (MFEBs) composed of $3\times 3$, $5\times 5$ and $7\times 7$ convolutional layers combined with global average pooling (GAP) to further extract features from $\bm F_{i}^{Fusion}$. The features at different scales in each MFEB are concatenated to obtain $\tilde{\bm F}_{2,i}^{Fusion}$ and $\tilde{\bm F}_{i}^{Fusion}$. Subsequently, they are processed through a feature extraction block. The results can be represented
as follows:
\begin{equation}
\begin{aligned}
    \bm F_{2,i}^{w}=Sigmoid\big( BN \big(Conv_{3\times 3}\big(\tilde{\bm F}_{2,i}^{Fusion}\big) \big) \big)
\end{aligned}
\end{equation}
\begin{equation}
\begin{aligned}
    \bm F_{i}^{w}=Sigmoid\big( BN \big(Conv_{3\times 3}\big(\tilde{\bm F}_{i}^{Fusion}\big) \big) \big)
\end{aligned}
\end{equation}
where $Sigmoid$ and $BN$ represent the sigmoid activation function and batch normalization layer, respectively. Based on $\bm F_{2,i}^{w}$ and $\bm F_{i}^{w}$, parameters for modulating $\bm F_{2,i}^{cat}$ and $\bm F_{i}^{cat}$ can be obtained:
\begin{equation}
\begin{aligned}
    \{\bm W_{2,i}, \bm W_{i}\}=Softmax\big( \big[ \bm F_{2,i}^{w}, \bm F_{i}^{w}\big] \big)
\end{aligned}
\end{equation}
where $\bm W_{2,i}(\cdot,\cdot,\cdot)+\bm W_{i}(\cdot,\cdot,\cdot)=1$ ($i=1,3$). Contrasting with traditional attention mechanisms that typically rely on spatial or channel attention, our unique modulation parameter prediction mechanism introduces a novel approach. It independently assigns specific weights to features at each spatial position, circumventing the need for conventional spatial and channel attention. This distinct capability allows for the targeted enhancement of features at individual spatial positions, offering a more refined and adaptable control over the feature processing compared to existing methods. The features modulated by $\bm W_{2,i}$ and $\bm W_{i}$ can be represented as:
\begin{equation}
\begin{aligned}
    \bm F_{2,i}=Conv_{1\times 1}\big(\bm W_{2,i} \odot \bm F_{2,i}^{Cat}+\bm W_{i} \odot \bm F_{i}^{Cat} \big)
\end{aligned}
\end{equation}

Following the aforementioned procedure, we can obtain the features $\bm{F}_{2,1}$ and $\bm{F}_{2,3}$, which are the results of the interaction and fusion between the reference image and the overexposed and underexposed images, respectively. These features are concatenated and subsequently processed through a $1\times 1$ convolution to generate the fused feature $\bm F_{M}^{0}$.

\subsubsection{Ghosting Suppression Module}
The FIFM effectively integrates the features from both the reference and non-reference images. However, owing to object movement in multi-exposure images, the direct utilization of the output features from the FIFM for HDR image reconstruction unavoidably gives rise to ghosting artifacts. To address this issue, taking into account the absence of ghosting artifacts in the SHDR results, this paper proposes a ghost artifact suppression method guided by SHDR. The method leverages the information disparity between the SHDR and multi-exposure HDR synthesis results to effectively suppress ghosting artifacts.
\begin{figure}[t!]
\centering
\includegraphics[width=2.6in,height=1.3in]{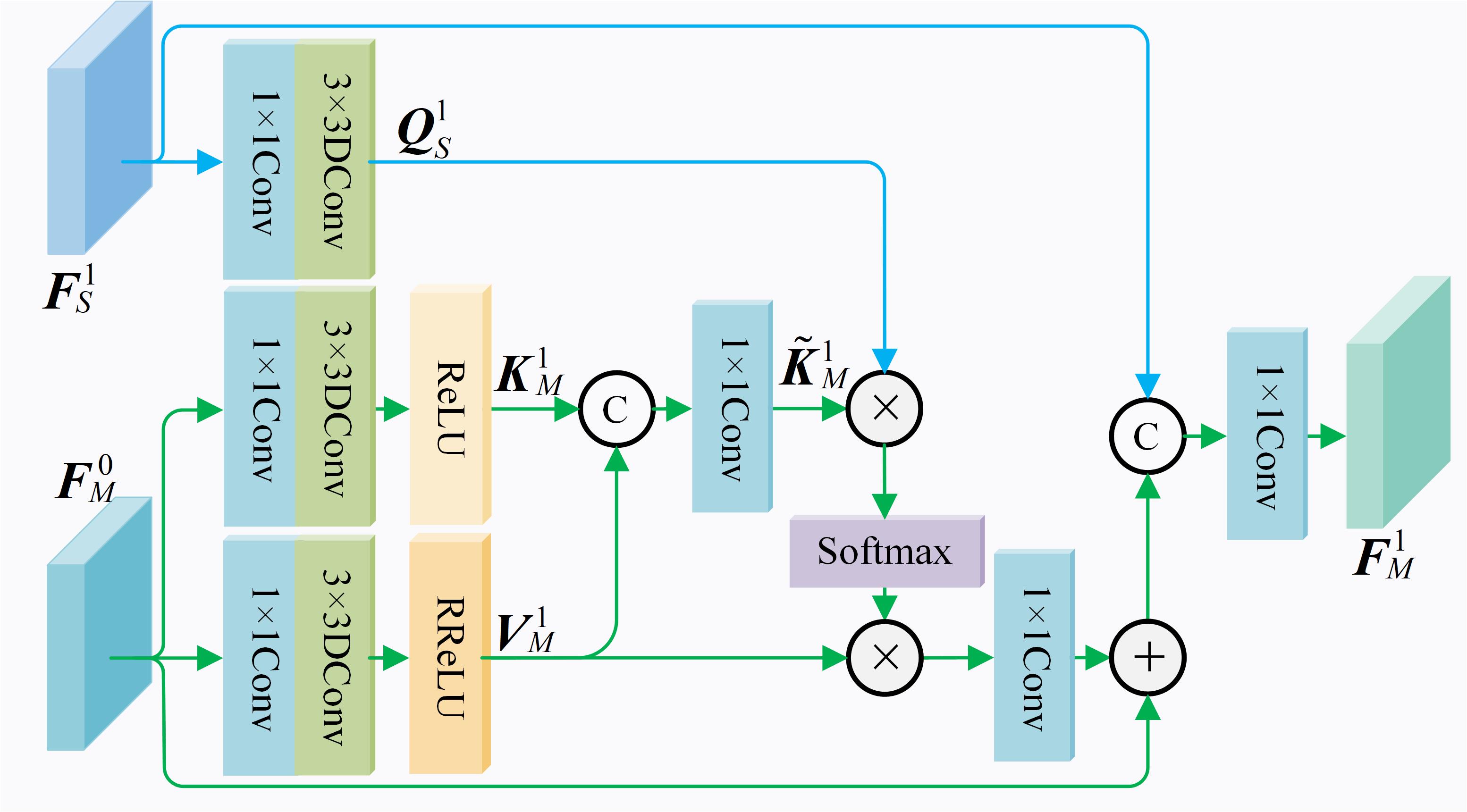}
\caption{Illustration of the ghost suppression module (GSM).}
\label{label6}
\end{figure}

Specifically, to fully utilize the valuable information in the fusion feature derived from FIFM, this paper designs a ghost suppression module (GSM) to suppress ghosting artifacts. This module adopts a cross-attention mechanism incorporating ReLU and RReLU activation functions, as shown in Fig.\ref{label6}. In the GSM, $\bm Q_{S}^{1}$ is obtained according to the following equation:
\begin{equation}
\begin{aligned}
    \bm Q_{S}^{1}=Conv_{3, 3}( Conv_{1\times 1}(\bm F_{S}^{1} ))
\end{aligned}
\end{equation}
where $Conv_{3, 3}$ represents a $3\times 3$ depthwise separable convolution. Correspondingly, the input feature $\bm F_{M}^{0}$ undergoes the similar operations to derive features $\bm K_{M}^{1}$ and $\bm V_{M}^{1}$:
\begin{equation}
\begin{aligned}
    \bm K_{M}^{1}=ReLU \big( Conv_{3, 3}\big( Conv_{1\times 1}\big(\bm F_{M}^{0} \big) \big) \big)
\end{aligned}
\end{equation}
\begin{equation}
\begin{aligned}
    \bm V_{M}^{1}=RReLU \big( Conv_{3, 3}\big( Conv_{1\times 1}\big(\bm F_{M}^{0} \big) \big) \big)
\end{aligned}
\end{equation}
where $RReLU=-min(0,x)$. Although the convolutional operation structure employed in $\bm K_{M}^{1}$ and $\bm V_{M}^{1}$ remains consistent, it's important to note that the parameters of the convolutional layers are not shared.

In Eqs. (20) and (21), to emphasize the significance of the ignored information in $\bm F_{M}^{0}$, this paper employs two different activation functions. Specifically, the $ReLU$ function activates features that contribute positively to enhancing reconstruction quality and possess amplitudes greater than 0, whereas the $RReLU$ function is responsible for activating features that positively influence reconstruction quality and exhibit amplitudes less than 0. In light of the complementary characteristics of features $\bm K_{M}^{1}$ and $\bm V_{M}^{1}$, we concatenate them and then process them through a $1\times 1$ convolution to achieve feature $\tilde{\bm K}_{M}^{1}$.

$\bm Q_{S}^{1}$ originates from SHDR-ESI, and it is inherently free from ghosting artifacts. Consequently, the shared information between $\bm Q_{S}^{1}$ and $\tilde{\bm K}_{M}^{1}$ contributes positively to the HDR image reconstruction, while the disparate information is regarded as containing ghosting artifacts. If the input feature can be adjusted based on the similarity between $\bm Q_{S}^{1}$ and $\tilde{\bm K}_{M}^{1}$, it becomes feasible to suppress ghosting artifacts while accentuating the information that positively contributes to the reconstruction process. To attain this objective, we employ the following method for feature purification:
\begin{equation}\small
\begin{aligned}
    \bm F_{M}^{1}=Conv_{1\times 1}\bigg[ Conv_{1\times 1}\bigg( Softmax\bigg( \frac{\bm Q_{S}^{1} (\tilde{\bm K}_{M}^{1})^{T}}{s} \bigg) \bm V_{M}^{1} \bigg)+\bm F_{M}^{0}, \bm F_{S}^{1} \bigg]
\end{aligned}
\end{equation}
where $s$ is a learnable scale parameter. $\bm F_{M}^{1}$ represents the output of the first GSM, and the output of the last GSM is denoted as $\bm F_{M}^{N}$. $\bm F_{M}^{N}$, along with the reference image feature $\bm F_{2}$, is fed into the reconstruction network to obtain the HDR image.

\subsection{Loss Function}
In practice, HDR images are typically exhibited following a tone mapping process. Therefore, training an HDR reconstruction network on tone-mapped images proves to be a more effective approach compared to direct training in the HDR domain. Given an image $\bm H$ in the HDR domain, we use $\mu$-law to perform tone mapping on $\bm H$:
\begin{equation}
\begin{aligned}
    T(\bm H)=\frac{\log(1+\mu \bm H)}{\log(1+\mu)}
\end{aligned}
\end{equation}
where $\mu$ is the compression factor, and $T(\bm H)$ represents the tone-mapped image. Following \cite{Kalantari-2017-Deep-scenes}, we set $\mu$ to 5000. 

In this paper, we use the following loss function to update the network parameters in an end-to-end manner:
\begin{equation}
\begin{aligned}
     L= L_{M}+\lambda L_{S}
\end{aligned}
\end{equation}
where $\lambda$ is a hyperparameter, $L_{M}$ represents the loss function for the SHDR-A-MHDR network, and $L_{S}$ represents the loss function for the SHDR-ESI network. $L_{M}$ and $L_{S}$ can be formulated as:
\begin{equation}
\begin{aligned}
    L_{l}= L_{l,re}+\alpha L_{l,ssim}+\beta L_{l,gradient}
\end{aligned}
\end{equation}
where $l=\{M,S\}$. $\alpha$ and $\beta$ are hyperparameters. $L_{l,re}$ is the reconstruction loss, $L_{l,ssim}$ is the structural similarity loss, and $L_{l,gradient}$ represents the gradient loss. The definition of $L_{l,re}$ in this paper is:
\begin{equation}
\begin{aligned}
    L_{l,re}=\big|\big|T(\bm H_{l})-T(\bm H_{G})\big|\big|_{1}
\end{aligned}
\end{equation}
where $\bm H_{S}$ and $\bm H_{M}$ represent the HDR images generated by SHDR-ESI and SHDR-A-MHDR, respectively, while $\bm H_{G}$ represents the ground truth. To ensure a high structural similarity between the generated HDR image and the ground truth, $L_{l,ssim}$ used in this paper is defined as follows:
\begin{equation}
\begin{aligned}
    L_{l,ssim}=1-SSIM(T(\bm H_{l}),T(\bm H_{G}))
\end{aligned}
\end{equation}

To ensure that the generated HDR image contains more edge details, this paper uses the Sobel operator to estimate the gradient maps of the generated image and the ground truth, then computes the $l_{1}$ loss on the estimated gradient map. Therefore, the $L_{l,gradient}$ in this paper is defined as follows:
\begin{equation}
\begin{aligned}
    L_{l,gradient}=\big|\big|\bigtriangledown (\bm H_{l})-\bigtriangledown (\bm H_{G})\big|\big|_{1}
\end{aligned}
\end{equation}
where $\bigtriangledown (\cdot)$ represents the Sobel operator.

\section{Experiments}
\subsection{Experimental Settings}
\textbf{Datasets}. To evaluate the effectiveness of the proposed method, we conduct performance validation on the Kalantari's\cite{Kalantari-2017-Deep-scenes}, Hu's\cite{Hu_2020_CVPR_Workshops}, Sen's\cite{sen2012robust} and Tursun's\cite{tursun2016objective} datasets, and compare its performance with several state-of-the-art methods. Kalantari's dataset contains 74 groups of training and 15 groups of test samples. Each sample contains three LDR images with exposure values of $\left\{-2, 0, +2\right\}$ or $\left\{-3, 0, +3\right\}$ respectively, and each set of images is accompanied by ground truth data. 
Hu's dataset consists of 100 samples, each comprising three LDR images with exposure values $\left\{-2, 0, +2\right\}$. Following \cite{Yan-2023-unified}, we designate 85 samples as the training set and the remaining 15 samples as the test set. Sen's and Tursun's datasets provide multi-exposure LDR images for 8 and 16 scenes, respectively, and neither of these datasets includes ground truth HDR images. We train the proposed network using the training set of Kalantari's dataset and subsequently tested it on the test set of Kalantari's dataset, as well as Sen's and Tursun's datasets. In addition, we train and test the model on Hu's dataset.

\textbf{Evaluation Metrics}. In order to quantitatively evaluate the quality of reconstruction results, PSNR-L, SSIM-L, PSNR-$\mu$, SSIM-$\mu$ and HDR-VDP-2\cite{mantiuk2011hdr, Eilertsen-2021-how} are used as objective evaluation metrics for the datasets with ground truth. For datasets without ground truth, BTMQI\cite{BTMQI}, MEF-SSIMd\cite{MEF-SSIMd} and UDQM\cite{tursun2016objective} are utilized. PSNR-L and SSIM-L evaluate the similarity between the reconstructed HDR image and the ground truth by calculating the peak signal-to-noise ratio (PSNR) and structure similarity index measure (SSIM) in the linear domain. PSNR-$\mu$ and SSIM-$\mu$ evaluate the quality of the reconstructed results by calculating the PSNR and SSIM between the images generated after $\mu$-law tone mapping and the ground truth images also processed with $\mu$-law tone mapping. HDR-VDP-2 assesses the visual quality of the reconstructed HDR image by predicting the difference between the reconstructed HDR image and the ground truth.
BTMQI assesses the overall quality of a tone-mapped HDR image by quantifying its information entropy, naturalness and structural information. A smaller BTMQI value indicates a higher reconstruction quality. MEF-SSIMd evaluates the comprehensive assessment of the HDR reconstruction image by individually measuring SSIM between it and corresponding sequences in the dynamic and static regions, followed by averaging the quality measurements from these two regions. UDQM evaluates the suppression effects on ghosting artifacts in reconstructed HDR images. This is achieved through the integration of blending metric ($Q_B$), gradient inconsistency metric ($Q_G$), visual difference metric ($Q_V$) and dynamic range metric ($Q_D$).

\begin{figure*}[t!]
\centering
\includegraphics[width=0.77\textwidth]{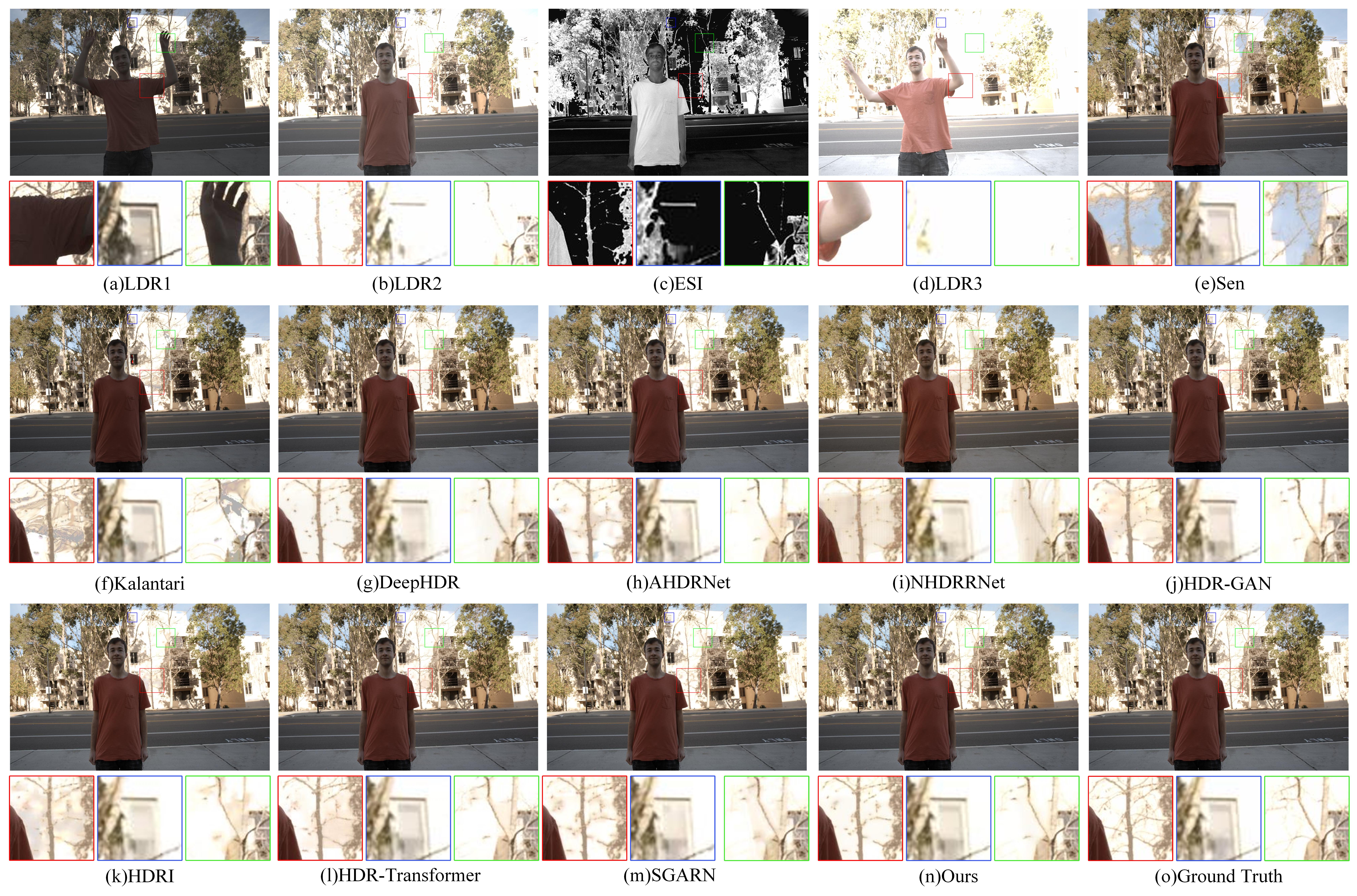}
\caption{Comparison of HDR reconstruction results by different methods on the ``Building" scene in the Kalantari’s test dataset.}
\label{label7}
\end{figure*}

\begin{figure*}[t!]
\centering
\includegraphics[width=0.77\textwidth]{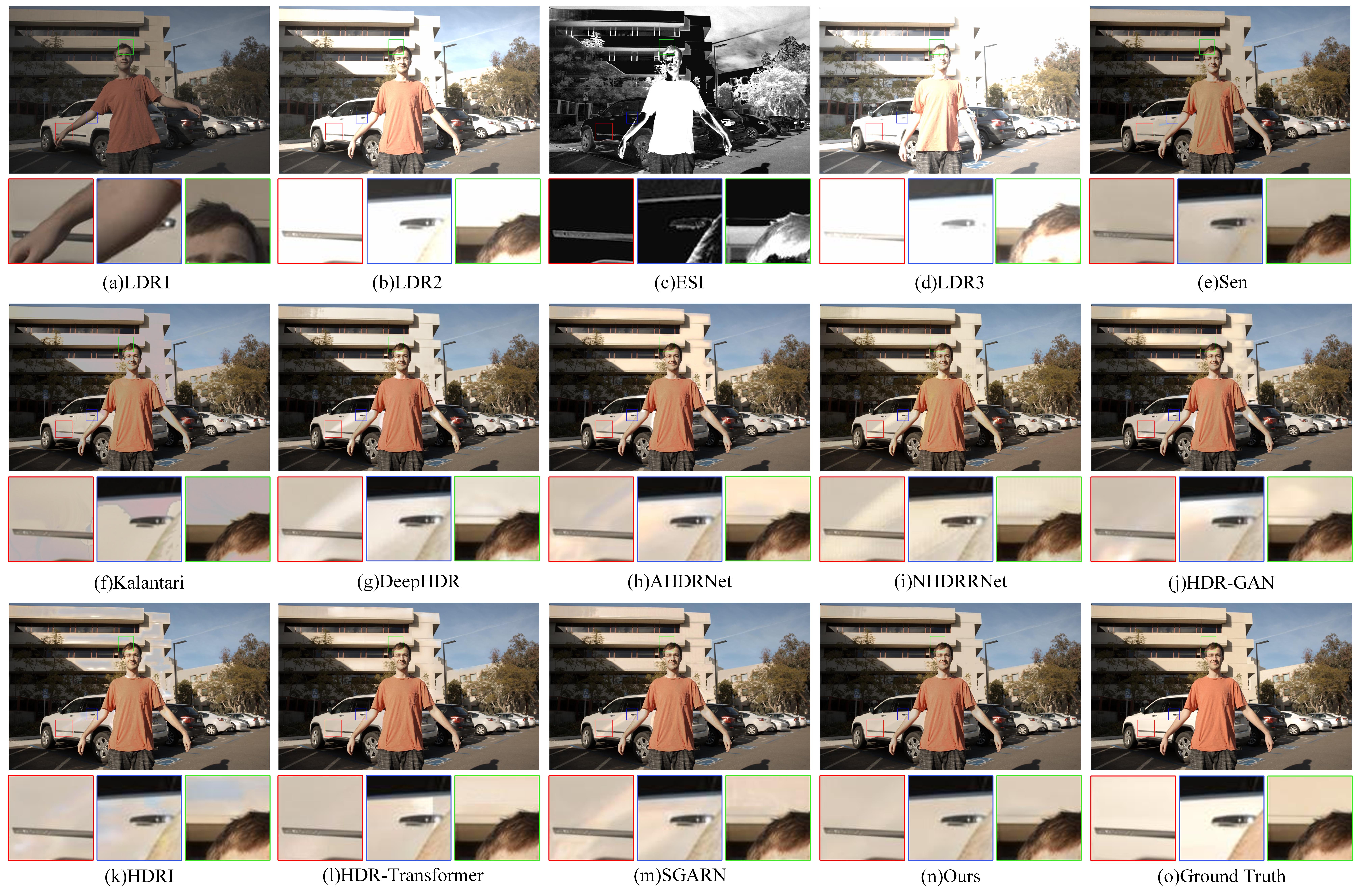}
\caption{Comparison of HDR reconstruction results by different methods on the ``Parking" scene in the Kalantari’s test dataset.}
\label{label8}
\end{figure*}

\textbf{Implementation Details}. The proposed method is implemented using the PyTorch framework and trained on a single NVIDIA 3090 GPU. In the experiments, the batch size is set to 2, and the model is trained for a total of 5000 epochs. In the training process, an Adam optimizer is used with an initial learning rate of 2e-4, weight decay set to 0, momentum values of $\beta_{1}$=0.9 and $\beta_{2}$=0.99, and $\epsilon$ set to 1e-8 for parameter updates. The learning rate is decayed by a factor of 0.1 every 2000 epochs.

\begin{figure*}[t!]
\centering
\includegraphics[width=0.77\textwidth]{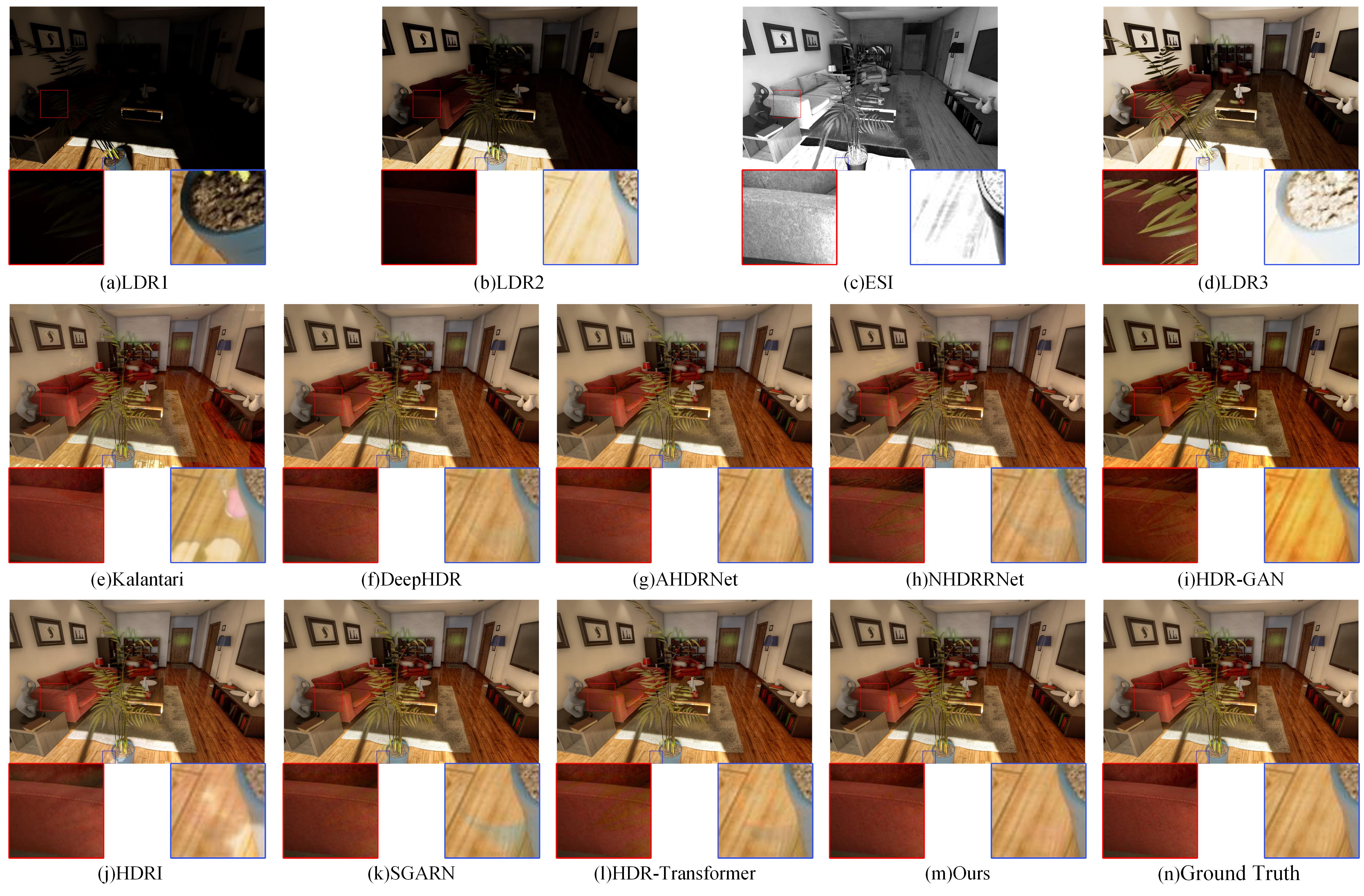}
\caption{Comparison of HDR reconstruction results by different methods on the ``089" scene in the Hu’s dataset.}
\label{label25}
\end{figure*}

\begin{table*}[!ht]
\centering {\caption{Quantitative evaluation of the reconstruction results on Kalantari's test set and Hu’s test set for different methods. Best and runner-up values are in bold and underlined, respectively.}\label{Table1}
\renewcommand\arraystretch{1.3}
\resizebox{\linewidth}{!}{
\begin{tabular}{c|c|c|c|c|c|c|c|c|c|c}
\hline
\multirow{2}*{methods}&\multicolumn{5}{c|}{Kalantari’s dataset}& \multicolumn{5}{c}{Hu’s dataset}\\
 \cline{2-11}
 &PSNR-$\mu$&SSIM-$\mu$ &PSNR-L&SSIM-L&HDR-VDP-2&PSNR-$\mu$&SSIM-$\mu$ &PSNR-L&SSIM-L&HDR-VDP-2\\
\hline
Sen\cite{sen2012robust}&40.95&0.9832&38.31&0.9753 &60.33&-&-&-&-&-\\
Kalantari\cite{Kalantari-2017-Deep-scenes}&42.74&0.9877&40.72&0.9824 &62.87
&41.60&0.9914&43.76&0.9938 &64.70\\
DeepHDR\cite{Wu-2018-Deep-motions}&41.62&0.9865&40.88&0.9858 &57.37 &44.70 	&0.9945 &44.27 &0.9960 & 68.90\\
AHDRNet\cite{Yan-2019-AHDRNet}&43.74&0.9913&41.69&0.9869 &63.51
&45.80 &\underline{0.9956} &46.51 &0.9985 & \textbf{70.65}\\
NHDRRNet\cite{Yan-2020-Non-local}&42.41&0.9887&41.08&0.9861 &62.31 &45.15 &0.9945 &48.75 &\underline{0.9989} &67.45\\
HDR-GAN\cite{Niu-2021-HDR-GAN}&43.92&0.9905&41.57&0.9865 &64.70 &45.86 &\underline{0.9956} &49.14 &0.9981 &61.12\\
HDRI\cite{Chung-2022-WACV}&43.65&0.9894&41.67&0.9867 &64.46 &43.77 &0.9930 &46.31 &0.9975 &67.82\\
HDR-Transformer\cite{Liu-2022-Ghost-free}&\underline{44.32}&\textbf{0.9916}&\underline{42.18}&\underline{0.9884} &64.63 &\underline{46.14} &\textbf{0.9961} &\underline{50.04} &0.9988 &68.92\\
SGARN\cite{tang2023structure}&43.96&0.9907&41.51&0.9874 &\underline{65.11} &45.26	&0.9944 &48.58 &0.9983 &\underline{70.18}\\
Ours&\textbf{44.39}&\underline{0.9915}&\textbf{42.20}&\textbf{0.9891} &\textbf{65.21} &\textbf{46.98} &\textbf{0.9961} &\textbf{52.49} &\textbf{0.9991} &69.76\\
\hline
\end{tabular}}}
\end{table*}

\subsection{Comparison with State-of-the-Art Methods}
To validate the effectiveness of the proposed method, we conduct experiments comparing it with several state-of-the-art methods on the Kalantari’s, Hu’s, Sen’s and Tursun’s datasets. These methods include Sen’s method\cite{sen2012robust}, Kalantari’s method\cite{Kalantari-2017-Deep-scenes}, DeepHDR\cite{Wu-2018-Deep-motions}, AHDRNet\cite{Yan-2019-AHDRNet}, NHDRRNet\cite{Yan-2020-Non-local}, HDR-GAN\cite{Niu-2021-HDR-GAN}, HDRI\cite{Chung-2022-WACV}, HDR-Transformer\cite{Liu-2022-Ghost-free} and SGARN\cite{tang2023structure}. Among these, Sen’s method is patch-based, while the rest are deep learning-based methods.

\begin{figure*}[t!]
\centering
\includegraphics[width=0.77\textwidth]{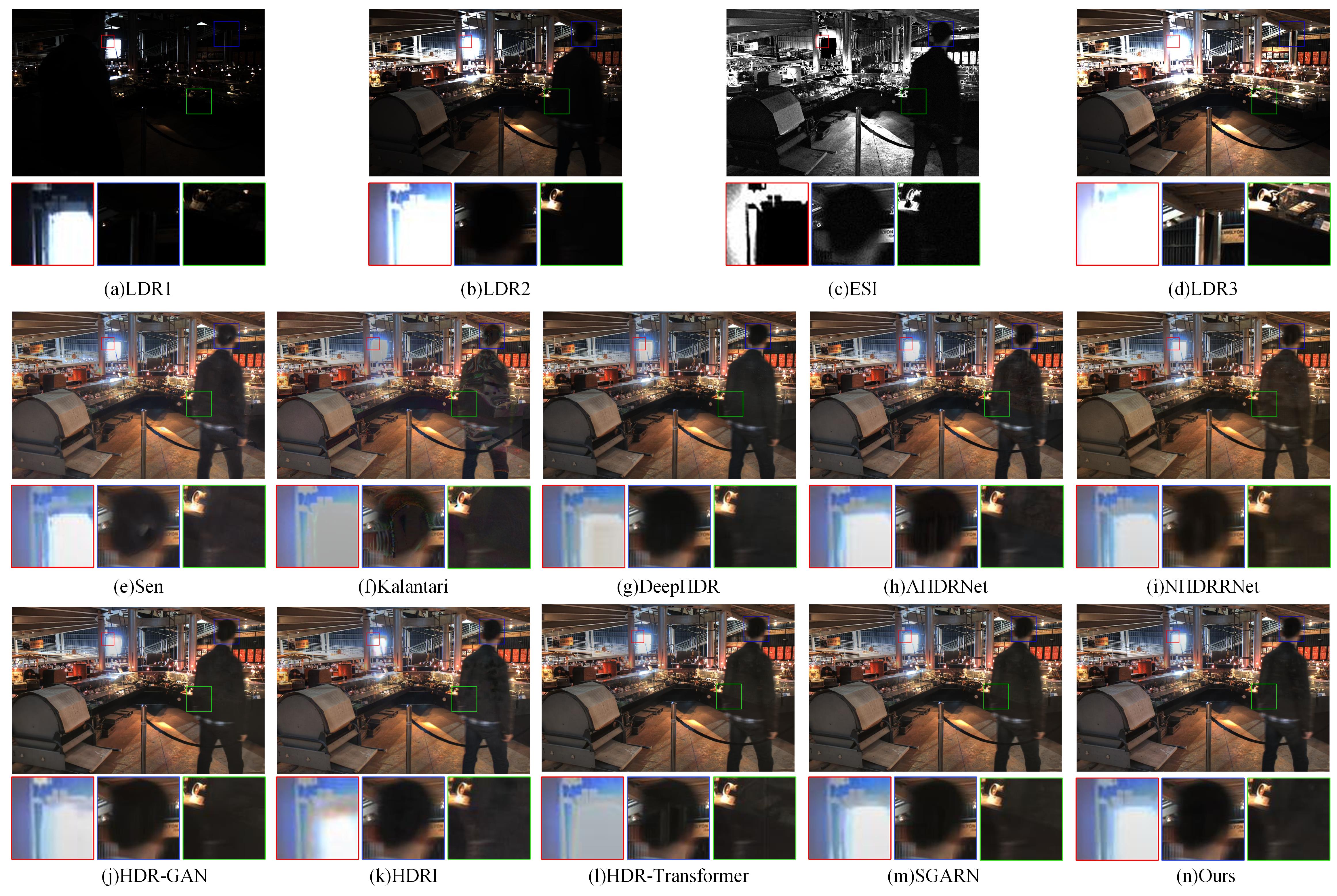}
\caption{Comparison of HDR reconstruction results by different methods on Sen's dataset.}
\label{label9}
\end{figure*}
\begin{figure*}[t!]
\centering
\includegraphics[width=0.77\textwidth]{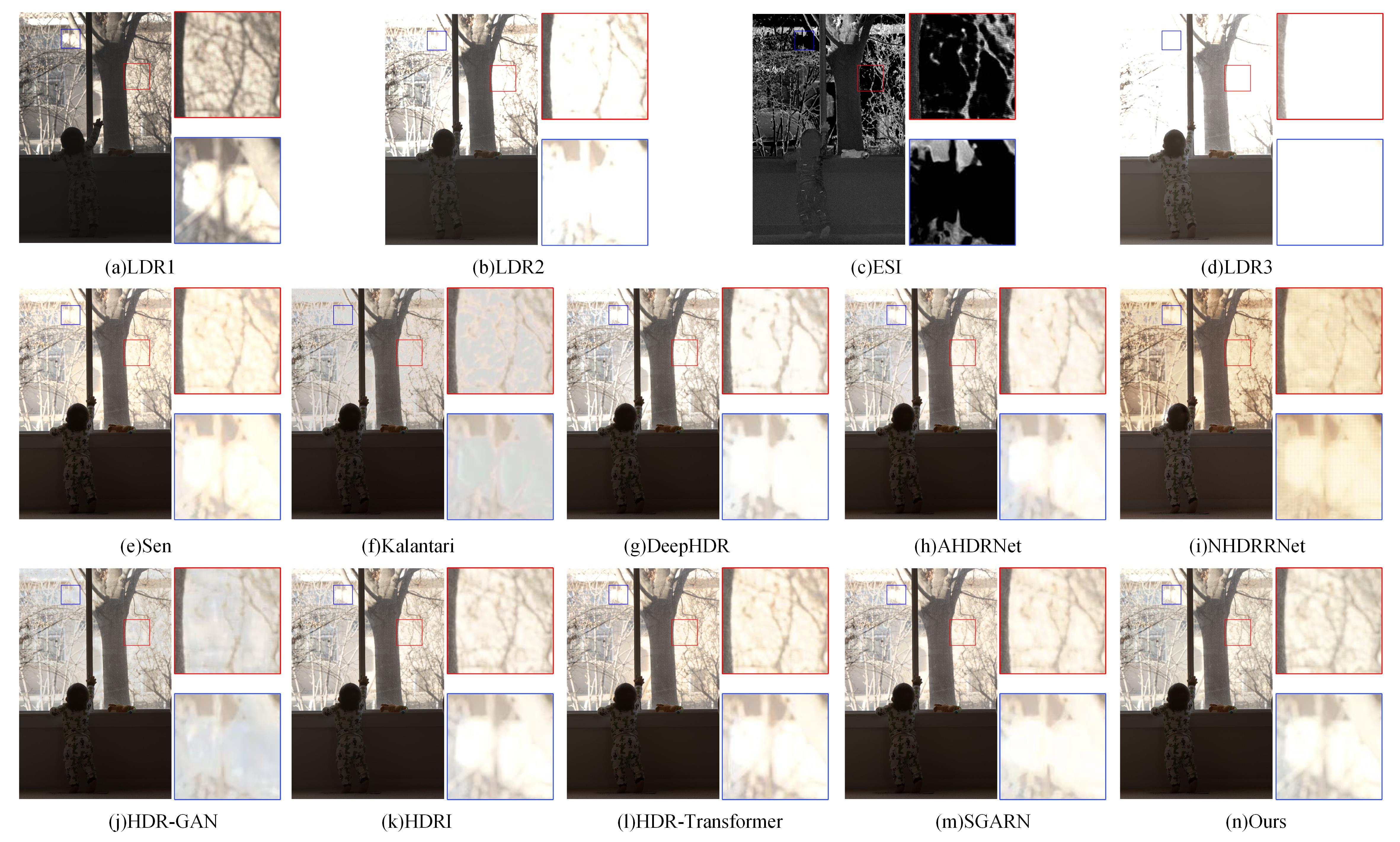}
\caption{Comparison of HDR reconstruction results by different methods on Tursun’s dataset.}
\label{label10}
\end{figure*}

\subsubsection{Experiments on Kalantari’s dataset}
To validate the effectiveness of our method, we initially compare the performance of different methods on the Kalantari's dataset. From the boxed region in Fig. \ref{label7}, it can be observed that Sen's method tends to make errors when searching for corresponding patches among images with different levels of saturation. This leads to a lower quality of reconstructed HDR images. Kalantari's method introduces ghosting artifacts in the reconstructed results due to inaccurate optical flow estimation. DeepHDR does not consider the recovery of missing details in saturated areas, resulting in significant detail loss in the generated HDR images. AHDRNet utilizes attention to suppress ghosting artifacts, but there is still room for significant improvement in the visual quality of the reconstructed HDR images. NHDRRNet and HDR-GAN do not consider the spatial inconsistency caused by moving objects, resulting in noticeable ghosting artifacts in the reconstructed results. HDRI and SGARN do not sufficiently address the recovery of details lost in overexposed areas, resulting in less clear details in the reconstructed HDR images. Moreover, due to the lack of an effective ghost artifact suppression mechanism, HDR-Transformer exhibits poor performance in ghost artifact suppression. In contrast, our method exhibits excellent performance in ghost artifact suppression and detail recovery. This is achieved through the incorporation of ESI to complement details in oversaturated regions and by utilizing the ghost-free SHDR results to effectively suppress potential ghost artifacts in the multi-exposure HDR synthesis results. Fig. \ref{label8} presents the reconstruction results of comparative methods on the Kalantari's test dataset in the Parking scene. Similar conclusions can also be drawn from the comparative results of Fig. \ref{label7}.

To objectively evaluate the reconstruction quality of different methods, Table \ref{Table1} presents the average objective evaluation results of different methods on the 15 test scenes from the Kalantari’s dataset. These results indicate that the proposed method achieves the best performance in terms of PSNR-$\mu$, PSNR-L, SSIM-L and HDR-VDP-2, and it also performs second best in terms of SSIM-$\mu$. This further confirms the effectiveness of the proposed method.

\begin{table}[!ht]\small
\centering {\caption{Quantitative evaluation of the reconstruction results on Sen’s dataset and Tursun’s dataset for different methods.}\label{Table3}
\renewcommand\arraystretch{1.3}
\resizebox{\linewidth}{!}{
\begin{tabular}{c|c|c|c|c|c}
\hline
 \multirow{2}*{methods}&\multicolumn{2}{c|}{Sen’s dataset}& \multicolumn{3}{c}{Tursun’s dataset}\\
 \cline{2-6}
 &BTMQI&MEF-SSIMd&BTMQI&MEF-SSIMd&UDQM\\
\hline
Sen\cite{sen2012robust}&3.7242&0.7896&4.2771&0.6557 &0.3938\\
Kalantari\cite{Kalantari-2017-Deep-scenes}&3.7679&0.7636&4.7089&0.6532&0.3984\\
DeepHDR\cite{Wu-2018-Deep-motions}&3.7744&0.7990&4.1628&0.6359&0.4147\\
AHDRNet\cite{Yan-2019-AHDRNet}&3.6877&0.7957&3.9205&0.6570&0.4260\\
NHDRRNet\cite{Yan-2020-Non-local}&3.8326&0.7908&3.9008&0.6278&0.3660\\
HDR-GAN\cite{Niu-2021-HDR-GAN}&7.1340&0.7992&7.0330&\underline{0.6663}&0.4118\\
HDRI\cite{Chung-2022-WACV}&5.5036&\underline{0.8024}&5.0423&0.6283&0.4098\\
HDR-Transformer\cite{Liu-2022-Ghost-free}&\underline{2.9057}&0.7956&\underline{3.4278}&0.6372&0.4142\\
SGARN\cite{tang2023structure}&3.6473&0.8003&3.7816&0.6612&\underline{0.4311}\\
Ours&\textbf{2.7677}&\textbf{0.8033}&\textbf{3.2496}&\textbf{0.6686} &\textbf{0.4329}\\
\hline
\end{tabular}}}
\end{table}

\subsubsection{Experiments on Hu’s dataset}
In Fig. \ref{label25}, a visual comparison on Hu's dataset between our proposed method and other state-of-the-art methods is presented. The comparison clearly shows that Kalantari's method tends to introduce color distortion in the reconstructed HDR images. While the results from DeepHDR and AHDRNet show potential, they still leave considerable room for improvement in visual quality. Both NHDRRNet and HDR-GAN exhibit notable artifacts in their reconstructed results, with HDR-GAN also displaying color distortion. The HDRI method, although effective, requires further enhancement in recovering content from saturated areas. SGARN and HDR-Transformer, on the other hand, perform less effectively in suppressing ghosting artifacts. In contrast, our proposed method excels in ghosting artifact suppression, demonstrating superior performance. Furthermore, Table \ref{Table1} lists the average objective evaluation results for various methods across the 15 testing scenes of Hu's dataset. These results clearly indicate that our method outperforms others in most metrics, thereby reinforcing the effectiveness of our proposed approach.

\subsubsection{Experiments on Sen’s dataset}
To further validate the effectiveness of the proposed method, we conduct a comparative analysis of the reconstruction performance of different methods on Sen's dataset. From the enlarged region in Fig. \ref{label9}, it is evident that the compared methods exhibited limited information recovery capabilities, resulting in the inefficient recovery of lost information in the overexposed areas. In contrast, the model proposed in this paper reconstructed HDR images with clearer details and minimal distortion, ultimately yielding optimal visual quality in the reconstructed results. The quantitative evaluation results presented in Table \ref{Table3} indicate that our proposed method exhibits superior performance compared to other methods, achieving the best results in both BTMQI and MEF-SSIMd. It should be noted that the Sen’s dataset does not provide the required camera parameters (ExposeTime, ISO, and F-Number) for calculating UDQM metric. Therefore, the performance of different methods on the Sen’s dataset is only evaluated on BTMQI and MEF-SSIMd metrics. This further highlights the advantages of the proposed approach on Sen's dataset.

\subsubsection{Experiments on Tursun’s dataset}
On Tursun's dataset, the visual comparison between the method proposed in this paper and other state-of-the-art methods is presented in Fig. \ref{label10}. As depicted in Fig. \ref{label10} (b), the reference image contains not only substantial underexposure regions but also exhibits notable large-scale foreground motion. The comparative methods, as observed from the reconstruction results, fail to effectively recover details and colors in underexposure regions, leading to distortions in both color and details. In the moving areas, the compared methods inevitably introduce information inconsistent with the reference image from underexposed and overexposed regions into the reconstruction results, creating ghosting artifacts in the heads and backs of the foreground objects. In contrast, the proposed method can generate more satisfactory details in the saturated regions, effectively integrate information from misaligned areas in the source images, and reduce the impact of ghosting artifacts on the reconstruction results.
Furthermore, the data in Table \ref{Table3} highlights that, when compared to other methods, our approach demonstrates superior performance with the highest values for BTMQI, MEF-SSIMd, and UDQM on Tursun's dataset. This provides additional support for the efficacy of the proposed method on Tursun's data.

\begin{figure*}[t!]
\centering
\includegraphics[width=0.68\textwidth]{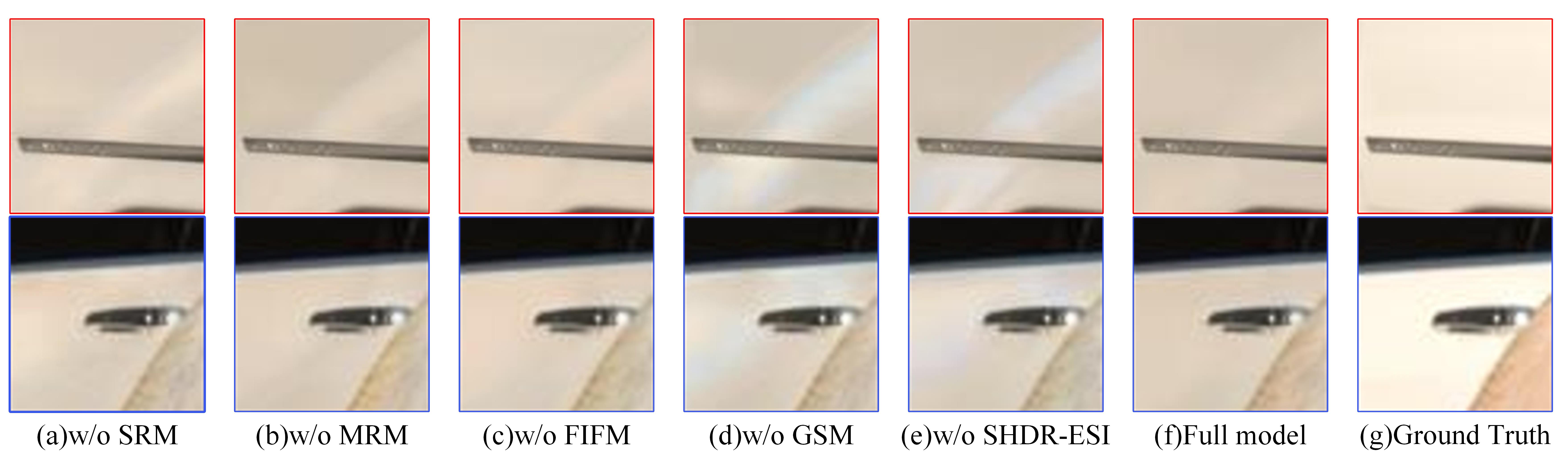}
\caption{Ablation experiments of different components.}
\label{label11}
\end{figure*}

\begin{figure*}[t!]
\centering
\includegraphics[width=0.8\textwidth]{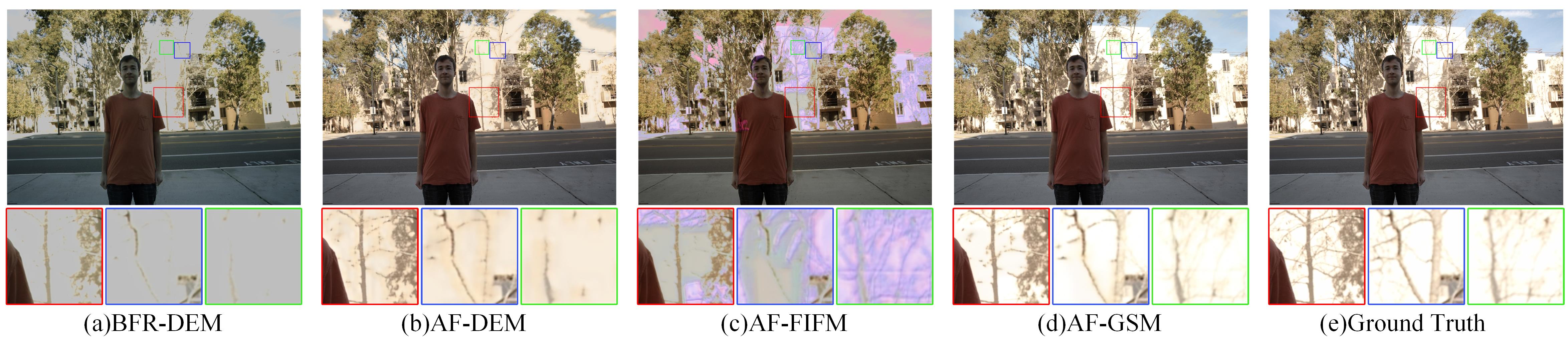}
\caption{Intermediate features reconstruction results of the crucial modules in the proposed method.}
\label{label12}
\end{figure*}

\subsection{Ablation Study}
The proposed network consists of SHDR-ESI branch and SHDR-A-MHDR branch. These two branches are principally constituted by the DEM, comprising SRM and MRM, along with FIFM and GSM. To evaluate the effectiveness of each module, the following ablation studies are conducted. During this procedure, considering the absence of an explicit baseline in our proposed method, we exclude the module or branch to be tested from the complete network structure to assess their respective contributions. Specifically, ``w/o SRM'' denotes the removal of SRM from the overall network structure. ``w/o MRM" signifies the exclusion of MRM from the overall network structure. ``w/o FIFM" entails the absence of FIFM in the overall network structure, instead directly concatenating features $\bm F_{1}$, $\bm F_{2}$ and $\bm F_{3}$ along the channel dimension before inputting them into the GSM module. ``w/o GSM" involves substituting the GSM module with direct channel-wise concatenation. Lastly, ``w/o SHDR-ESI" means removing the SHDR-ESI branch from the overall network structure. The interactive fusion result ($\bm F_{M}^{0}$) obtained from FIFM is directly fed into the reconstruction network for HDR image reconstruction. Fig. \ref{label11} displays local results of reconstructed HDR images using scene images shown in Fig. \ref{label8}. 

Additionally, to further validate the efficacy of the crucial modules in this study, features from both before and after each module are used to reconstruct images, respectively. These results provides a more intuitive demonstration of the effectiveness of these critical modules. Specifically, BFR-DEM refers to the preliminary stage preceding the execution of the DEM, whereas AF-DEM represents the subsequent phase that occurs after the execution of DEM. AF-FIFM denotes the phase that follows the execution of the FIFM, while AF-GSM describes the stage subsequent to the execution of the GSM. Fig. \ref{label12} demonstrates the reconstruction results of intermediate features of the crucial modules based on the scene images depicted in Fig. \ref{label7}.

\begin{table}[!ht]\small
\centering {\caption{
Objective evaluation of the effectiveness of each proposed module.}\label{Table2}
\renewcommand\arraystretch{1.3}
\resizebox{\linewidth}{!}{
\begin{tabular}{c|c|c|c|c|c}
\hline
&PSNR-$\mu$&SSIM-$\mu$ &PSNR-L&SSIM-L&HDR-VDP-2\\
\hline
w/o SRM&44.17&0.9914&41.84&0.9884 &65.05\\
w/o MRM&44.18&0.9914&41.88&0.9885 &64.95\\
w/o FIFM&44.14&0.9914&41.65&0.9883 &64.80\\
w/o GSM&44.09&0.9912&41.69&0.9878 &65.02\\
w/o SHDR-ESI&43.98&0.9909&41.73&0.9885 &\textbf{65.43}\\
Full model&\textbf{44.39}&\textbf{0.9915}&\textbf{42.20}&\textbf{0.9891} &65.21\\
\hline
\end{tabular}}}
\end{table}

\textbf{Effectiveness of SRM:} The SRM in this paper is primarily utilized to emphasize features that have a positive effect on HDR reconstruction, thus aiding in the effective recovery of detailed information. Initially, the SRM is excluded from the comprehensive model, followed by a comparative analysis to assess the differences between its results and the performance of the full model. From Fig. \ref{label11}(a), it can be observed that the model without SRM yields HDR images with blurred details at the boundaries and slight color distortion. 
This observation is further corroborated by the objective evaluation results presented in presented in Table \ref{Table2}, which show a decrease in overall performance when SRM is removed. Specifically, PSNR-$\mu$ and PSNR-L decrease by 0.22dB and 0.36dB, respectively. This validates the positive role of SRM in enhancing detailed information in the reconstructed results.

\textbf{Effectiveness of MRM:} The proposed MRM compensates for the lack of detail information in overexposed regions by transferring information from ESI to the reference image. From Fig. \ref{label11}(b), it is evident that in the absence of MRM, the model fails to reconstruct HDR images with clear details, and its ability to suppress ghosting artifacts decreases compared to the full model. Furthermore, as evident from Table \ref{Table2}, the removal of MRM results in a decrease in all evaluated performance metrics compared to the full model. Specifically, PSNR-$\mu$ and PSNR-L exhibit reductions of 0.21 dB and 0.32 dB, respectively. This provides additional evidence of the effectiveness of MRM in enhancing the quality of HDR image reconstruction.

As shown in Fig. \ref{label12}, the BFR-DEM image displays an oversaturated region that notably lacks detailed information. In contrast, the corresponding region in the AF-DEM image reveals a richness of detail. This observation further underscores the effectiveness of the DEM, comprising SRM and MRM as proposed in this study, in highlighting and integrating crucial information from ESI and reference images, thereby enhancing the quality of HDR image reconstruction.

\textbf{Effectiveness of FIFM:} FIFM enhances the quality of HDR images by effectively integrating features from both the reference and non-reference images. As can be seen from Fig. \ref{label11}(c), when FIFM is removed from the overall model, the reconstructed HDR image appears blurry and ghosting artifacts. The objective evaluation results shown in Table \ref{Table2} indicate that, compared to the performance of the full model, omitting FIFM results in the significant reductions in both PSNR-L and HDR-VDP-2, with decreases of 0.55dB and 0.41 respectively. This demonstrates the essential role of FIFM in enhancing the overall performance of the model. Furthermore, the AF-FIFM image in Fig. \ref{label12} demonstrates FIFM’s proficiency in integrating features from both reference and non-reference images. However, it should be noted that this integration process also introduces ghosting artifacts within the fused features.

\textbf{Effectiveness of GSM:} To mitigate potential ghosting artifacts in the fused features resulting from FIFM, this paper introduces a GSM that is guided by the intermediate features of SHDR-ESI. This method effectively suppresses inconsistent information and accentuates key information by calculating the correlation between the intermediate features of SHDR-ESI and the fused features.
As observed in Fig. \ref{label11}(d), when GSM is removed from the overall model, the reconstructed HDR image exhibits very prominent ghosting artifacts. 
The objective evaluation data in Table \ref{Table2} indicates a notable decline in all metrics upon the exclusion of GSM. PSNR-$\mu$ and PSNR-L exhibits reductions of 0.30dB and 0.51dB, respectively, while SSIM-L decreases from 0.9891 to 0.9878. A comparison between the AF-FIFM image and the AF-GSM image in Fig. \ref{label12} further demonstrates the effectiveness of GSM in ghosting suppression.

\textbf{Effectiveness of SHDR-ESI:} After removing SHDR-ESI, the SHDR-A-MHDR no longer utilizes ESI for feature aggregation and highlighting essential information. Without SHDR-ESI, the interactively fused features are directly fed into the reconstruction network to generate HDR images. As shown in Fig. \ref{label11}(e), the SHDR-ESI branch plays a significant role in ghosting artifact suppression and detail recovery. Additionally, the results in Table \ref{Table2} indicate that in the absence of SHDR-ESI, except for HDR-VDP-2, all other the evaluation metrics exhibit a decline in performance. This demonstrates the effectiveness of SHDR-ESI.

\subsection{Further Discussion}
\begin{figure}[t!]
	\centering
	\includegraphics[width=3.4in,height=2.7in]{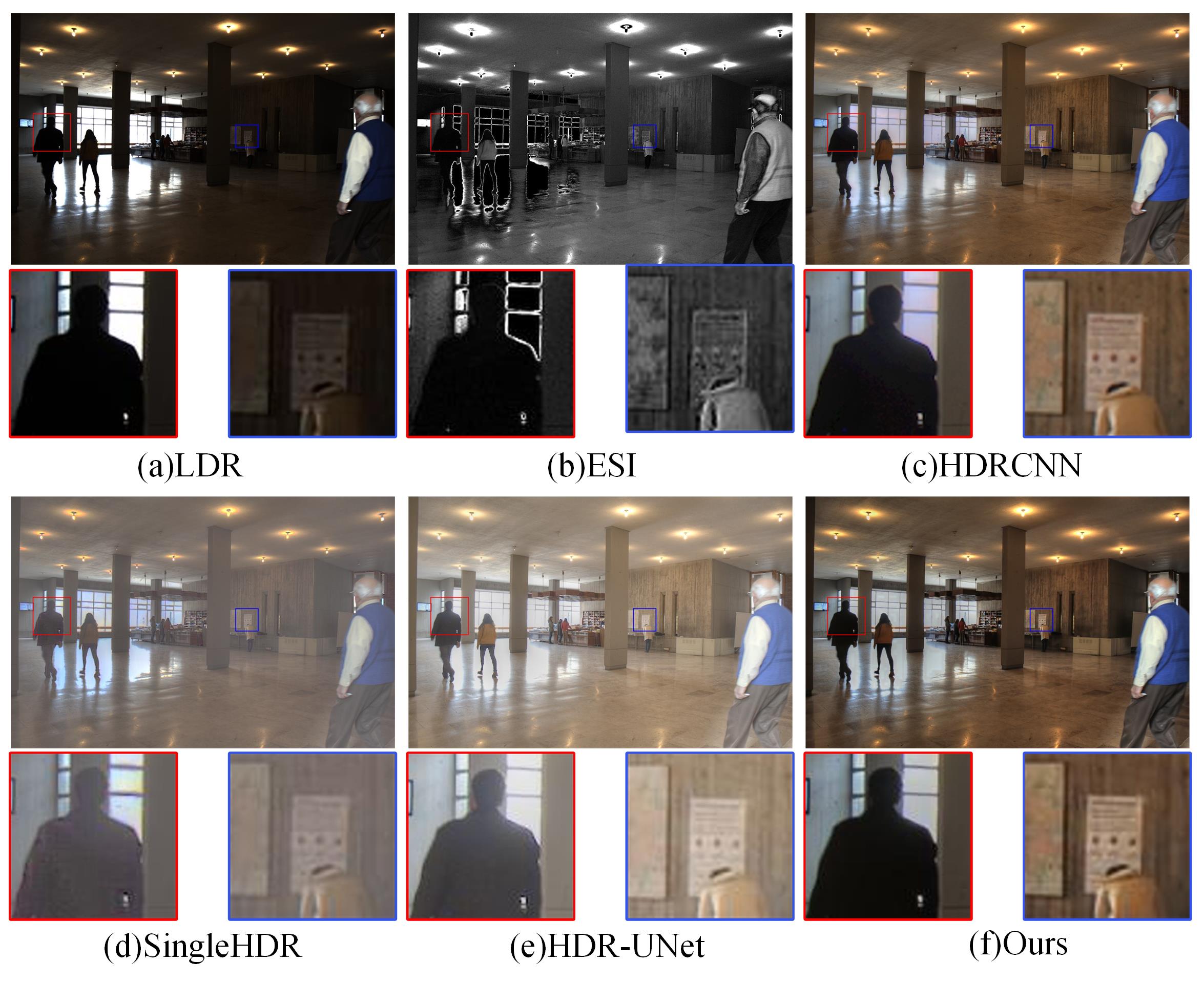}
	\caption{Comparison of HDR reconstruction results by different SHDR methods on the ``161" scene in Tursun's dataset.}
	\label{label31}
\end{figure}

To validate the effectiveness of the proposed SHDR-ESI branch, we evaluate its reconstruction results on the Tursun’s dataset and compare it with the existing SHDR methods, including HDRCNN\cite{Eilertsen-2017-HDR-CNNS}, SingleHDR\cite{Liu-2020-Single-pipeline} and HDR-UNet\cite{chen2021hdrunet}. The experiment results are depicted in Fig. \ref{label31} and Table \ref{Table31}. As can be seen in Fig. \ref{label31}, the HDR reconstruction images generated by SingleHDR and HDR-UNet exhibit low contrast and blurred details. In contrast, our method and HDRCNN have achieved commendable reconstruction results. As shown in Table \ref{Table31}, our method has achieved the best values on MEF-SSIMd and UDQM, while also obtaining comparable scores on BTMQI. Overall, the proposed SHDR-ESI branch, which highlights and aggregates crucial information in both ESI and the reference image, demonstrates superior capabilities in HDR image reconstruction.

\begin{table}[!ht]\small
	\centering {\caption{Quantitative evaluation of the reconstruction results from Tursun's dataset using various SHDR methods.}\label{Table31}
		\renewcommand\arraystretch{1.3}
		\resizebox{0.7\linewidth}{!}{
			\begin{tabular}{c|c|c|c}
				\hline
				methods&BTMQI&MEF-SSIMd&UDQM\\
				\hline
				HDRCNN\cite{Eilertsen-2017-HDR-CNNS}&\textbf{3.0431} &0.5536 &\underline{0.4192}\\
				SingleHDR\cite{Liu-2020-Single-pipeline}&4.1468 &\underline{0.6043} &0.3809 \\
				HDRUNet\cite{chen2021hdrunet}&3.5120 &0.5541 &0.4038 \\
				Ours&\underline{3.0475}&\textbf{0.6133}&\textbf{0.4333}\\
				\hline
	\end{tabular}}}
\end{table}

\begin{table*}[!ht]
	\centering {\caption{Performance comparison of the different methods on the number of model parameters , FLOPs and inference time.}\label{Table4}
		\renewcommand\arraystretch{1.3}
		\resizebox{\linewidth}{!}{
			\begin{tabular}{c|c|c|c|c|c|c|c|c|c|c}
				\hline
				methods &Sen\cite{sen2012robust} &Kalantari\cite{Kalantari-2017-Deep-scenes} &DeepHDR\cite{Wu-2018-Deep-motions} &AHDRNet\cite{Yan-2019-AHDRNet} &NHDRRNet\cite{Yan-2020-Non-local} &HDR-GAN\cite{Niu-2021-HDR-GAN} &HDRI\cite{Chung-2022-WACV} &HDR-Transformer\cite{Liu-2022-Ghost-free} &SGARN\cite{tang2023structure} &Ours\\
				\hline
				Size(M)& -&0.30	&16.61 &1.44 &41.42 &7.67 &6.70 &1.46 &0.89 &4.76\\
				FLOPs(G)& -& -& 15.93 &23.70 &5.11 &- &- &23.89 &13.76 &77.49\\
				Time (s)&61.81 &29.14 &0.28 &0.30 &0.31 &0.69& 20.61 &4.92 &0.53 &1.97\\
				\hline
	\end{tabular}}}
\end{table*}

\subsection{Parameter analysis}
The hyperparameters involved in this study include the number $N$ of DEMs and GSMs, $\lambda$ in Eq. (24), and $\alpha$ and $\beta$ in Eq. (25). In the proposed network, $N$ DEMs and $N$ GSMs are used to aggregate essential information and suppress potential ghosting artifacts. $\lambda$ is employed to balance the contribution of the loss function in the SHDR-ESI. $\alpha$ and $\beta$ are used to harmonize the contributions of the structural similarity loss and gradient loss, respectively. In the process of hyperparameter selection, we utilize Kalantari's test dataset to analyze how the model performance changes with different values of the aforementioned parameters.

\begin{figure}[t]
	\centering
	\vspace{-0.15in}
	\begin{minipage}{1\linewidth}	
		\subfigure[]{
			\label{fig:1}
			\includegraphics[width=0.46\linewidth,height=0.9in]{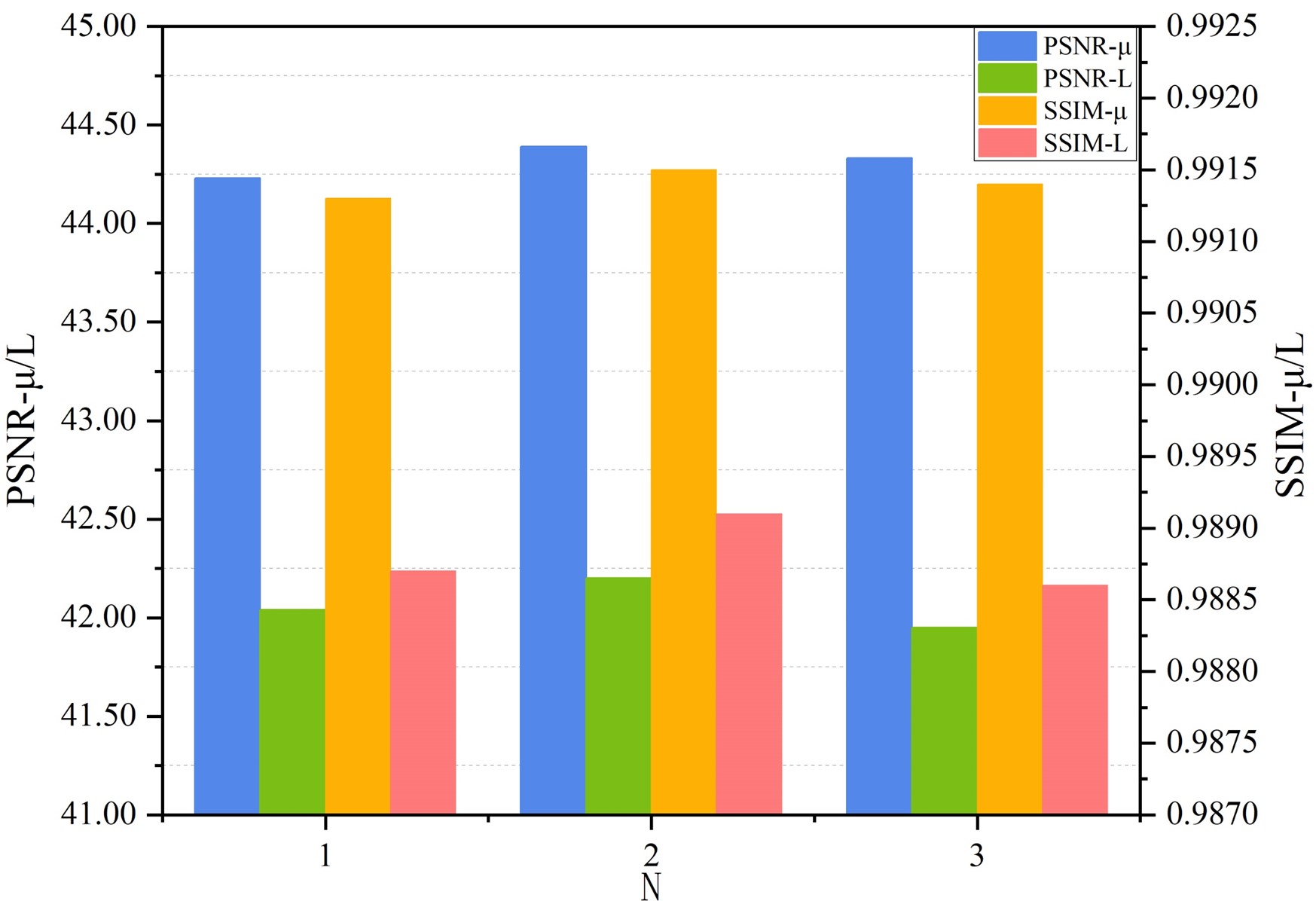}	
		}\noindent
		\subfigure[]{
			\label{fig:2}
			\includegraphics[width=0.46\linewidth,height=0.9in]{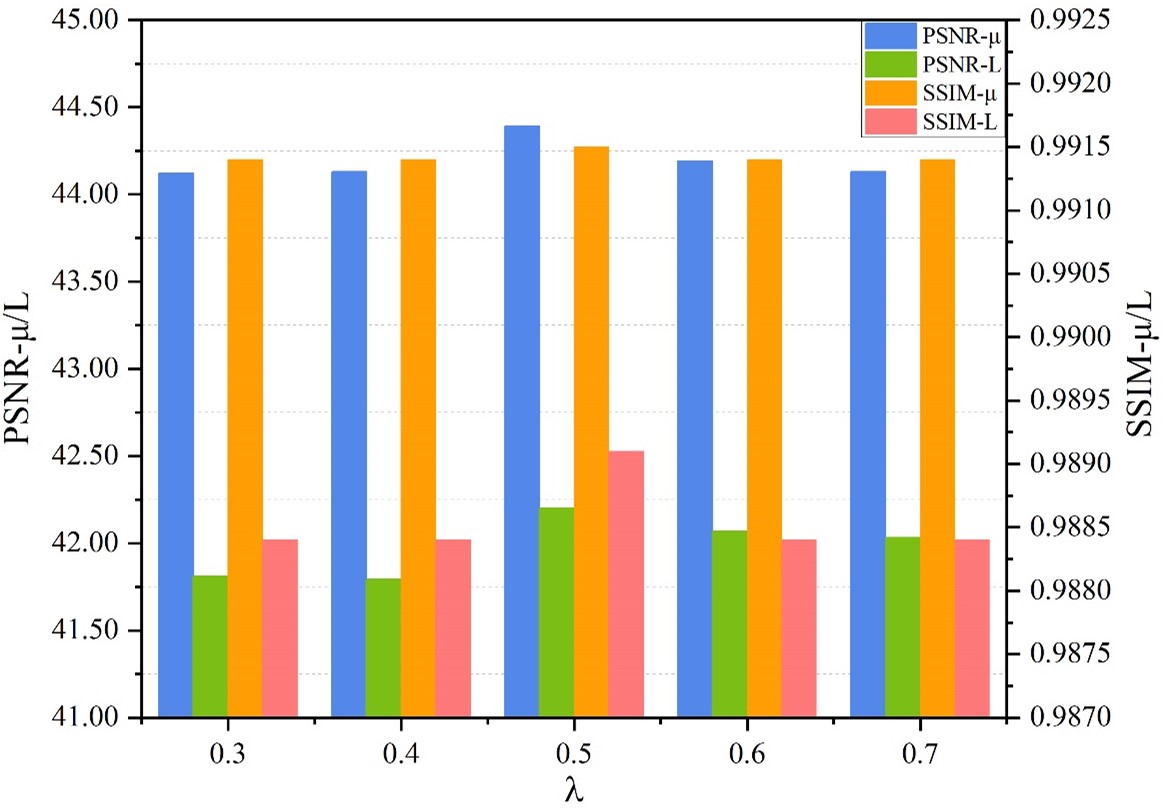}
		}
	\end{minipage}
	\vskip -0.3cm 
	\begin{minipage}{1\linewidth }
		\subfigure[]{
			\label{fig:3}
			\includegraphics[width=0.46\linewidth,height=0.9in]{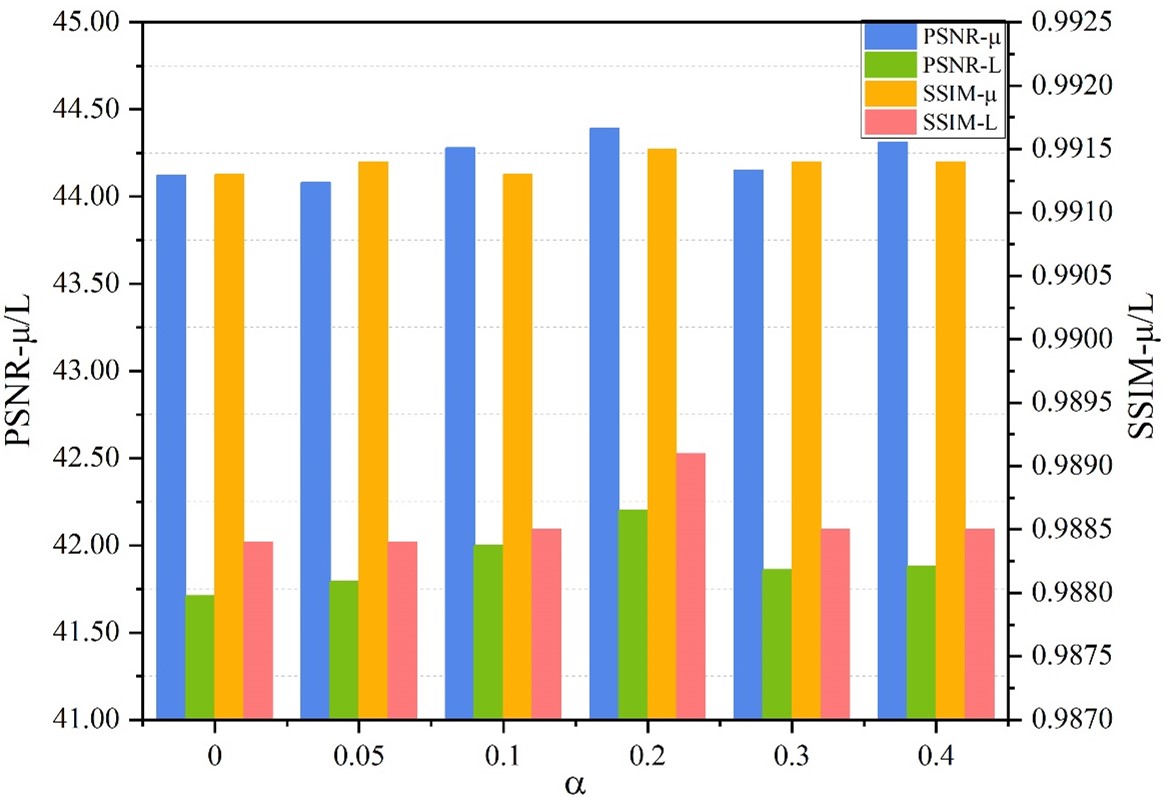}
		}\noindent
		\subfigure[]{
			\label{fig:4}
			\includegraphics[width=0.46\linewidth,height=0.9in]{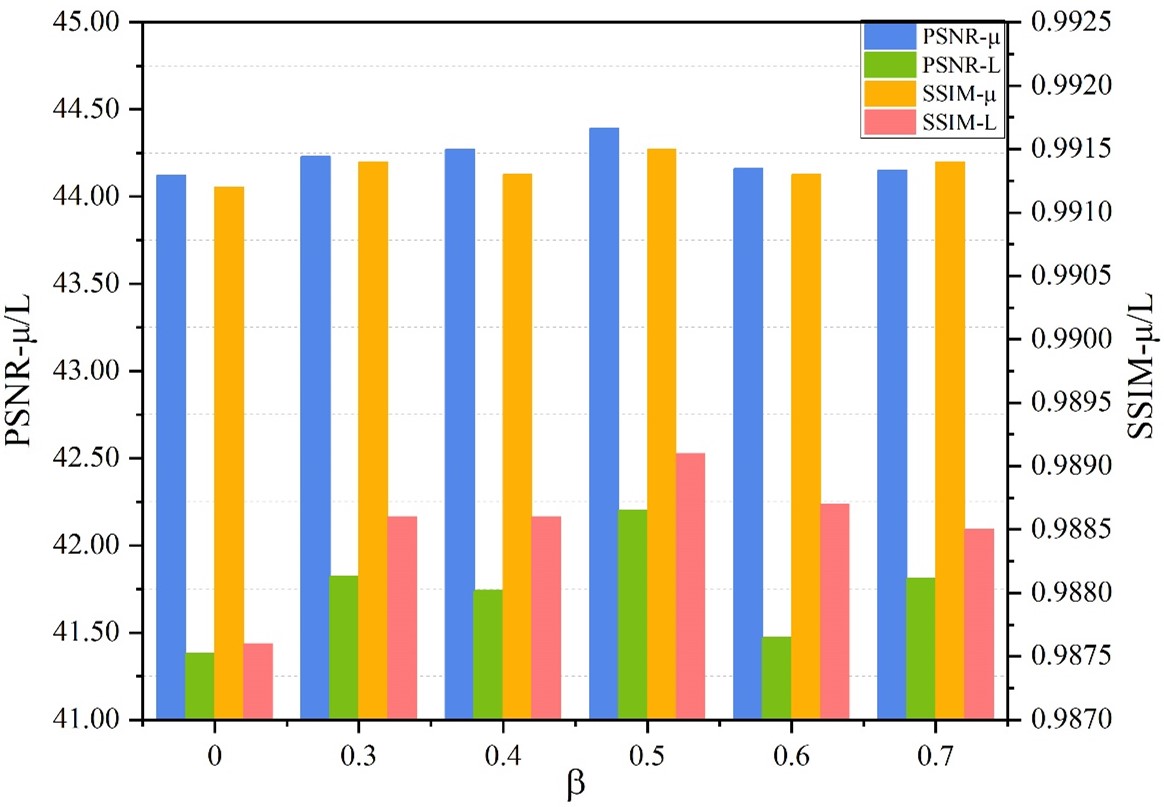}
		}
	\end{minipage}
	\vspace{-0.18in}
	\caption{Hyperparameter analysis on Kalantari's test dataset. (a), (b), (c) and (d) respectively represent the impact of variations in N, $\lambda$, $\alpha$ and $\beta$ on the model's performance.}
	\vspace{-0.2in}		
	\label{fig:1234}
\end{figure}

\textbf{The Influence of $N$ on Model Performance:} 
Fig. \ref{fig:1} illustrates the impact of different values of $N$ on the model's performance. It can be observed that the model achieves its best performance when $N=2$. Therefore, in this paper, $N$ is set to 2.

\textbf{The Influence of $\lambda$ on Model Performance:} Fig. \ref{fig:2} illustrates the curve of model performance as the parameter $\lambda$ varies when $\alpha$=0.2 and $\beta$=0.5. From the graph, it can be observed that the model achieves its best performance when $\lambda$=0.5. Therefore, in this study, the parameter $\lambda$ is set to 0.5.

\textbf{The Influence of $\alpha$ on Model Performance:} To assess the impact of the parameter $\alpha$ on model performance, this study keeps the parameters $\lambda$ and $\beta$ fixed at 0.5. Fig. \ref{fig:3} presents the variation in model performance with different values of $\alpha$. It can be observed that the model achieves its best performance when $\alpha$=0.2. Therefore, $\alpha$ is set to 0.2 in this study.

\textbf{The Influence of $\beta$ on Model Performance:} Fig. \ref{fig:4} illustrates the performance variation of the reconstruction results with different values of $\beta$ when $\lambda$=0.5 and $\alpha$=0.2. It can be observed that when $\beta$=0.5, the evaluation metrics all reach their best values. Therefore, we set $\beta$ to 0.5.

\subsection{Computational Complexity Analysis}
To evaluate the efficiency of all comparison methods, we analyze their computational complexity from three aspects: the number of model parameters, FLOPs, and inference time. The results are presented in Table \ref{Table4}. The inference time is calculated as the average time taken to process 15 scene images from Kalantari's test dataset. For FLOPs calculation, the size of the input images is standardized to 128$\times$128 pixels. As indicated in Table \ref{Table4}, our model has a relatively high FLOPs value, due to the incorporation of a single-frame image reconstruction model and blocks with specific functions. Nevertheless, the parameter size of our model is only 4.76M, suggesting that it is not excessively large. Moreover, the average inference time for our model to process a scene is 1.97 seconds, demonstrating that our model's inference speed is relatively fast and acceptable. Overall, our model can be practically deployed and applied after training. In future work, we aim to optimize the proposed model to reduce its FLOPs and improve training speed.

\section{Conclusion}
In this paper, we introduce a dual-branch network consisting of single-frame HDR image reconstruction and multi-exposure HDR image reconstruction. Within the unified framework, we achieve the suppression of ghosting artifacts in SHDR-A-MHDR and also restore missing details. In this process, the introduction of ESI effectively promotes the recovery and preservation of details in overexposed regions. In SHDR-A-MHDR, it mainly consists of two modules: FIFM and GSM. The FIFM thoroughly integrates features from both the reference and non-reference images, ensuring the richness of feature information while highlighting the features that play a positive role in the reconstructed image. Furthermore, GSM is primarily achieved through guidance from SHDR-ESI results. This process not only effectively suppresses the impact of ghosting on the fusion result but also emphasizes the contribution of consistent information to the reconstruction result, further enhancing the quality of the HDR image. The proposed method exhibits superior performance compared to other comparative methods on three publicly available datasets.

\bibliography{mybibfile}

\begin{IEEEbiography}[{\includegraphics[width=1in,height=1.25in,clip,keepaspectratio]{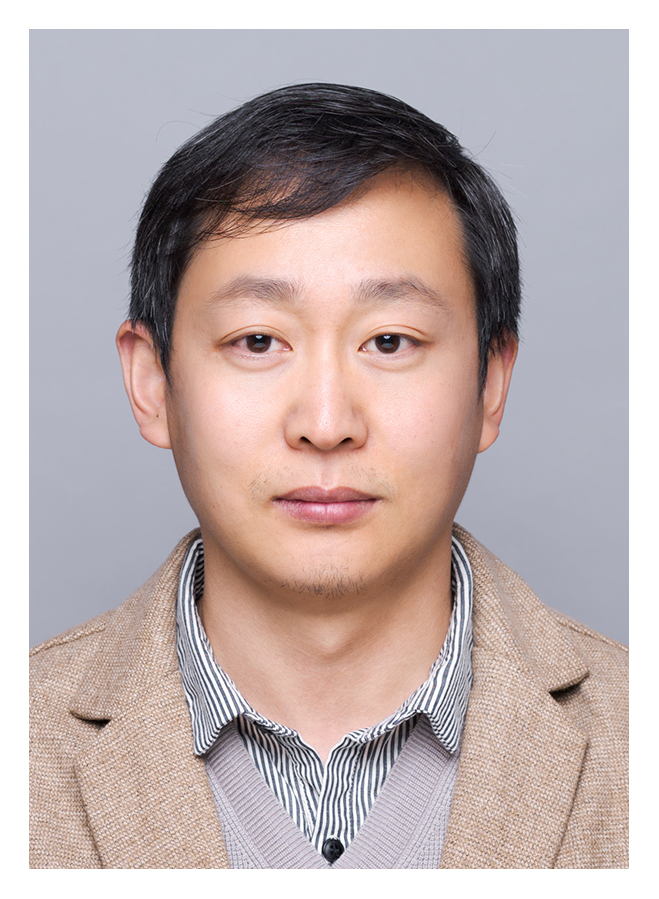}}]{Huafeng Li} received the M.S. degrees in applied mathematics major from Chongqing University in 2009 and obtained his Ph.D. degree in control theory and control engineering major from Chongqing University in 2012.  He is currently a professor at the School of Information Engineering and Automation, Kunming University of Science and Technology, China. His research interests include image processing, computer vision, and information fusion.
\end{IEEEbiography}

\begin{IEEEbiography}[{\includegraphics[width=1in,height=1.25in,clip,keepaspectratio]{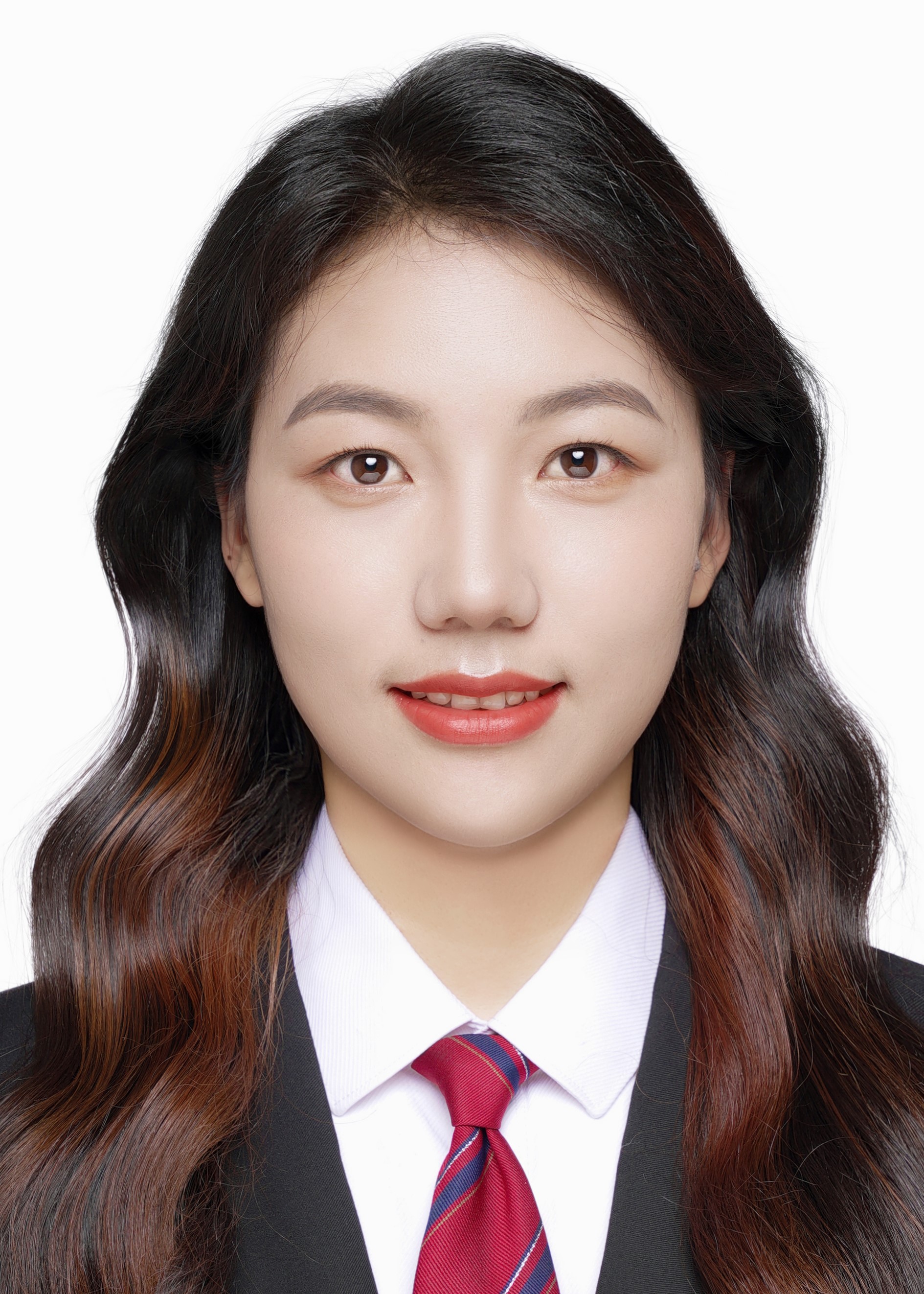}}]{Zhenmei Yang} received the B.S. degree from the Faculty of Information Engineering and Automation, Kunming University of Science and Technology, Kunming, China, in 2018. She is currently pursuing the master’s degree with the Kunming University of Science and Technology, Kunming, China. Her current research interests include computer vision and image processing.
\end{IEEEbiography}
	
\begin{IEEEbiography}[{\includegraphics[width=1in,height=1.25in,clip,keepaspectratio]{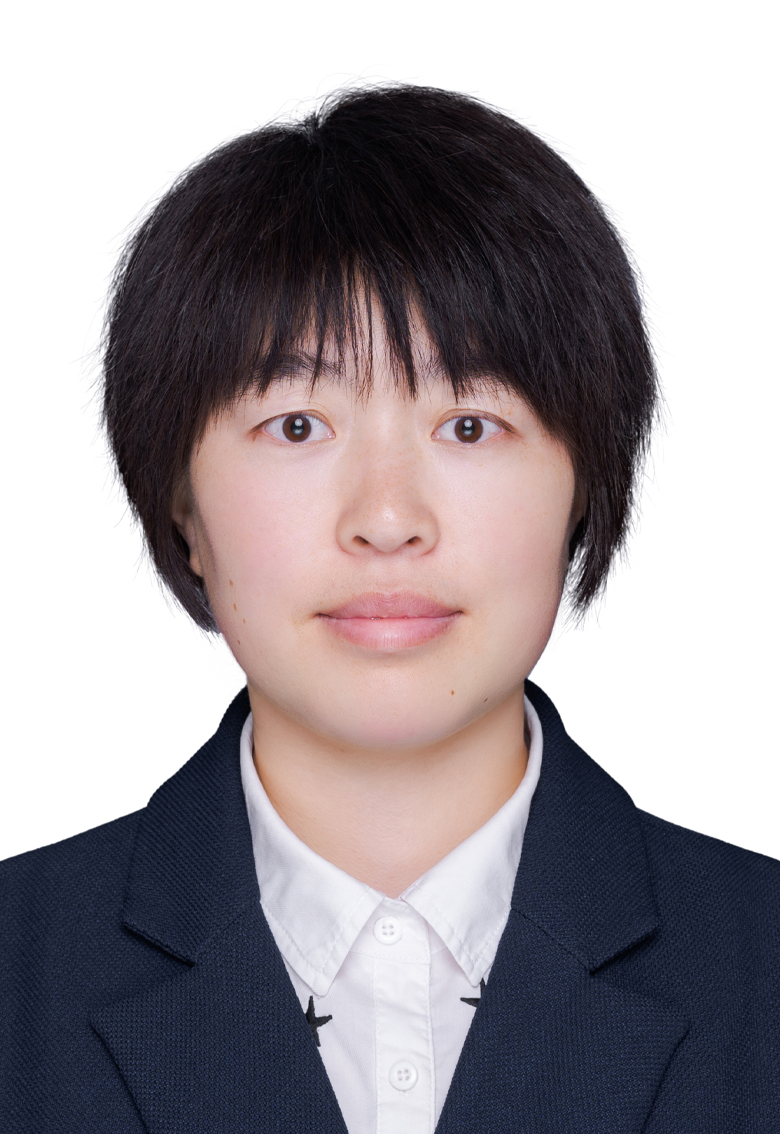}}]{Yafei Zhang} received the Ph.D. degree in signal and information processing from Institute of electronics, Chinese Academy of Sciences, Beijing, China, in 2008. She is currently an associate professor at College of Information Engineering and Automation, Kunming University of Science and Technology, Kunming, China. Her main research interests include image processing and pattern recognition.
\end{IEEEbiography}

\begin{IEEEbiography}[{\includegraphics[width=1in,height=1.25in,clip,keepaspectratio]{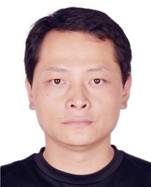}}]{Dapeng Tao} received the B.E. degree from Northwestern Polytechnical University, Xi’an, China, in 1999, and the Ph.D. degree from the South China University of Technology, Guangzhou, China, in 2014. He is currently a Professor with the School of Information Science and Engineering, Yunnan University, Kunming, China. He has authored or coauthored more than 50 scientific articles. He has served for more than ten international journals including IEEE TIP, IEEE TNNLS, IEEE TCYB, IEEE TMM, IEEE CSVT, Pattern Recognition, and Information Sciences. His research interests include machine learning, computer vision, and robotics. 
\end{IEEEbiography}

\begin{IEEEbiography}[{\includegraphics[width=1in,height=1.25in,clip,keepaspectratio]{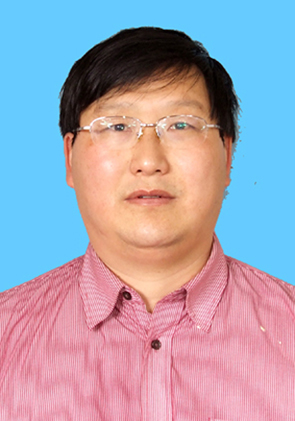}}]{Zhengtao Yu} received his Ph.D degree in computer application technology from Beijing Institute of Technology, Beijing, China, in 2005. He is currently a professor with the School of Information Engineering and Automation, Kunming University of Science and Technology, China. His main research interests include natural language process, image processing and machine learning.
\end{IEEEbiography}

\end{document}